\documentclass[prx,aps,showpacs,twocolumn, preprintnumbers,
amsmath,amssymb,superscriptaddress,nofootinbib,longbibliography,nofootinbibjustification=justified,singlelinecheck=false]{revtex4-2}

\usepackage{physics}
\usepackage{braket}
\usepackage{bm}
\usepackage{graphicx}
\usepackage{booktabs}
\usepackage{xspace}
\usepackage{multirow}
\usepackage{mathtools}

\usepackage{lipsum}
\usepackage{hyperref} 
\usepackage{cleveref}
\usepackage{listings}
\usepackage{color}
\usepackage[x11names,svgnames]{xcolor}
\usepackage{tcolorbox} 

\hypersetup{
    hidelinks,
    colorlinks=true,
    breaklinks=true,
    citecolor=SteelBlue,
    filecolor=LimeGreen,
    linkcolor=MediumBlue,
    urlcolor=MediumPurple
}

\newcommand{\mpl}{Max Planck Institute for the Science of Light, Staudtstrasse 2, 91058 Erlangen, Germany}
\newcommand{\tuebingen}{Machine Learning in Science Cluster, Department of Computer Science, Faculty of Science, University of Tübingen, Maria-von-Linden-Straße 6, 72076 Tübingen, Germany}
\newcommand{\jena}{Institute of Condensed Matter Theory and Optics, riedrich Schiller University Jena, Fröbelstieg 1, 07743 Jena}

\begin{document}

\title{Generating Interesting Scientific Ideas using Knowledge Graphs and LLMs: Evaluations with 100 Research Group Leaders}
 
\author{Xuemei Gu}
\email{xuemei.gu@outlook.com}
\affiliation{\mpl}
\affiliation{\jena}

\author{Mario Krenn}
\email{mario.krenn@uni-tuebingen.de}
\affiliation{\mpl}
\affiliation{\tuebingen}

\begin{abstract}
The rapid growth of scientific literature makes it increasingly challenging for researchers to identify novel and impactful ideas, especially across disciplines. Modern artificial intelligence (AI) systems offer new opportunities for scientific ideation, but how compelling are AI-generated ideas, and how can their quality be improved? Here, we introduce \textsc{SciMuse}, which generates personalized research ideas using a knowledge graph of 58 million papers and a large language model (LLM). A central focus of this work is to understand how interesting these ideas are. Therefore, we conducted a large-scale evaluation in which more than 100 research group leaders -- spanning the natural sciences to the humanities -- rated over 4,400 personalized ideas according to their level of interest. \textcolor{black}{Overall, expert ratings were modest (mean 2.40 on a 5-point scale, most common rating 1), while 24.9\% of ideas were rated 4 or 5. We find that supplying concept pairs selected using the knowledge graph does not improve expert-rated interest over a titles-only GPT baseline. High-citation-predicted pairs even showed a weak tendency (1.94$\sigma$) toward lower interest than random pairs. Nevertheless, graph features can be used to control properties of ideas, and, using this unique evaluation dataset, we show that idea interest can be predicted with both a supervised neural network based on graph features and a zero-shot ranking approach based on an LLM.} Our work provides an AI methodology for generating scientific ideas and a large-scale interdisciplinary expert evaluation, paving the way to study and improve difficult-to-measure metrics such as expert-perceived scientific interestingness. 
\end{abstract}

\maketitle

An interesting idea is often at the heart of successful research projects, crucial for their success and impact. However, with the accelerating growth in the number of scientific papers published each year \cite{fortunato2018science, wang2021science, bornmann2021growth}, it becomes increasingly difficult for researchers to uncover novel and interesting ideas. This challenge is even more pronounced for those seeking interdisciplinary collaborations, who have to navigate an overwhelming volume of literature.

Automated systems that extract insights from millions of scientific papers present a promising solution \cite{evans2011metaknowledge, wang2021science, lin2023sciscinet}. Recent advances have demonstrated that analyzing relationships between research topics across vast scientific literature can reliably predict future research directions \cite{rzhetsky2015choosing, krenn2020predicting, sybrandt2020agatha, nadkarni2021scientific, krenn2023forecasting}, forecast the potential impact of emerging work \cite{shi2023surprising, gu2025forecasting}, and identify unconventional avenues for discovery \cite{sourati2023accelerating}. With the advent of powerful large-language models (LLMs), it is now possible to leverage knowledge from millions of papers to generate concrete research ideas. Examples involve \textsc{SciMON}, a LLM system that is optimized for increasing the novelty of ideas \cite{wang2023scimon}, \textsc{ResearchAgent}, a system to iteratively create ideas in the domain of computer science \cite{baek2024researchagent}, \textsc{MOOSE} which is designed for automated open-domain scientific hypothesis discovery \cite{yang2023large}, \textsc{SciPiP} which generates ideas using structured literature databases including citation information \cite{wang2024scipip}, \textsc{IdeaSynth} which evolves idea primitives using literature-based feedback \cite{pu2025ideasynth}, or human-in-the-loop idea generators such as \textsc{SciIdeator} \cite{radensky2024scideator}.

Yet, a crucial question remains: Are AI-generated research ideas compelling to experienced scientists? While automated evaluations of research idea quality (for example, originality and feasibility) have recently been demonstrated, such as in \textsc{LiveIdeaBench} \cite{ruan2026evaluating}, human evaluation of AI-generated ideas by experienced scientists across diverse fields is still largely lacking. Previous studies have conducted small-scale evaluations with six natural language processing (NLP) PhD students \cite{wang2023scimon}, three social science PhD students \cite{yang2023large} and ten PhD students in computer science and biomedicine \cite{baek2024researchagent}. In recent studies on NLP research ideation, AI-generated ideas were rated as more novel than human-generated ones by 79 reviewers, mostly PhD students in computer science \cite{si2025can}, yet less executable according to a separate study of 43 participants \cite{si2025ideation}. However, perspectives from experienced researchers -- who define and evaluate research projects through grant applications and shape their group's research agenda -- are essential for assessing the value of new ideas. Involving a larger group of more senior evaluators across diverse scientific fields could provide deeper insights into what makes a research idea compelling, and how to generate and predict such ideas.

\begin{figure}[!t]
    \centering
    \includegraphics[width=1\linewidth]{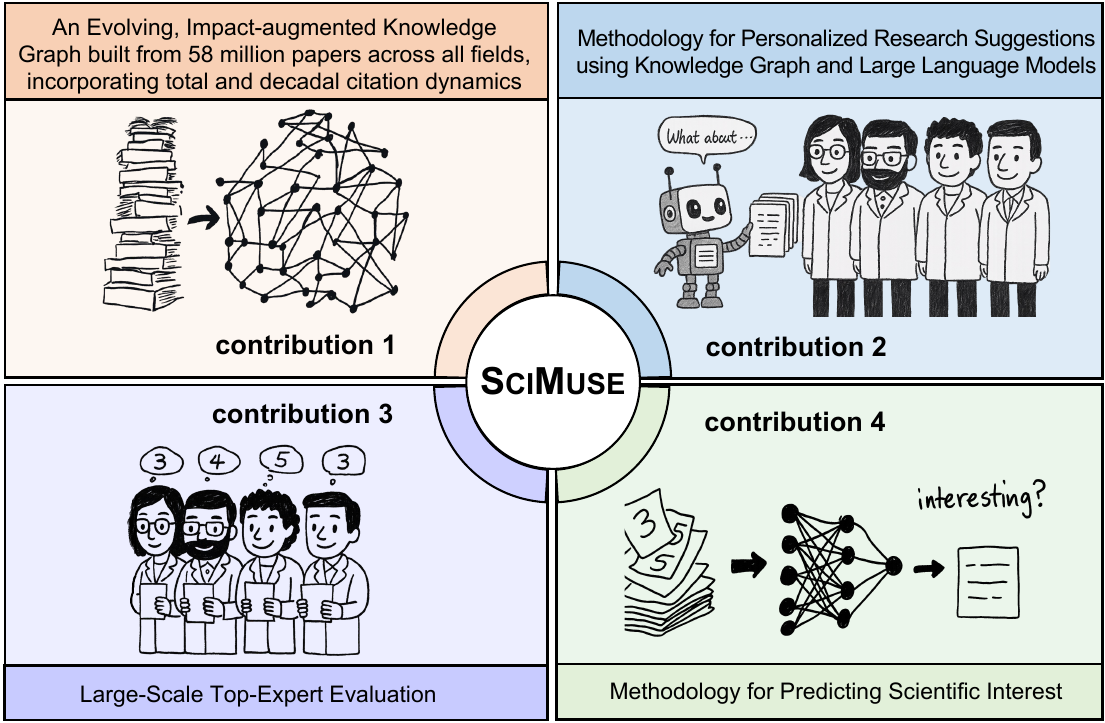}
    \caption{Overview of the four main components of \textsc{SciMuse}: an evolving impact-augmented knowledge graph, a method for generating personalized research suggestions, a large-scale expert evaluation across diverse scientific fields, and methods for predicting the scientific interest of research suggestions.}
    \label{fig:overview}
\end{figure}

Here, we introduce \textsc{SciMuse}, a system designed to suggest personalized research ideas for individual scientists and potential collaborations, and use it to conduct a large-scale expert evaluation of AI-generated ideas. As illustrated in Fig.~\ref{fig:overview}, the work combines four main components. First, it builds an evolving knowledge graph that captures scientific content and its impact from more than 58 million papers across a broad range of disciplines. On this basis, scientists’ research directions are represented as subgraphs, which are used together with LLMs to generate personalized collaboration suggestions and concrete research ideas. These ideas are then assessed in a large-scale expert evaluation, in which more than 100 Max Planck Society research group leaders from diverse disciplines rated over 4,400 personalized ideas on a five-point scale, yielding a unique expert-annotated dataset on scientific interest. Based on this evaluation dataset, we then develop models to predict the interest level of AI-generated ideas using both structured knowledge-graph information through supervised learning and unstructured text through LLM-based zero-shot ranking, thereby supporting the automated discovery of promising research directions.

Our work enables a large-scale, interdisciplinary expert evaluation of AI-generated research ideas to study scientific interest -- an inherently subjective, still underexplored, yet important dimension -- and to derive feedback signals for improving future models. Such evaluation-driven improvements could help future systems uncover opportunities that may not be readily apparent and strengthen the role of AI as a valuable source of inspiration in scientific discovery \cite{krenn2022scientific, hope2023computational, wang2023scientific, ai4science2023impact, musslick2025automating}.

\subsection*{Scientific Interest} 
While scientific interest is a critical factor in shaping the choice and pursuit of scientific directions, it remains unclear what \textit{interest} actually is, what function it serves, and what makes something interesting. In psychology, interest is often described as an \textit{epistemic emotion} (or knowledge emotion), alongside surprise, confusion, and awe \cite{silvia2006exploring, silvia2008interest}. Its function is commonly understood as providing intrinsic motivation for learning and exploration. Previous psychological studies of written texts have identified two main classes of predictors of interestingness: first, a combination of novelty and complexity; and second, a cluster of coherence-related variables, such as coherence and concreteness \cite{sadoski2001resolving, silvia2006exploring}. In our study, a scientist rating an idea as \textit{interesting} should therefore not be interpreted to mean that they would necessarily pursue it immediately, invest substantial resources in it, or regard it as the most important or feasible next project. Rather, we interpret interest as an early-stage signal in scientific discovery: the extent to which a proposal attracts sufficient curiosity and epistemic attention to merit further thought, discussion, or exploration, before later-stage considerations such as timing, resources, and strategic priorities come into play. In this sense, interest is related to, but distinct from, novelty, importance, feasibility, and uniqueness: an idea can be novel yet too obscure to be engaging, important yet obvious, feasible yet unexciting, or interesting yet currently impractical. This interpretation also resonates with recent science-of-science work showing that atypical or surprising combinations are often associated with increased scientific impact \cite{uzzi2013atypical,wang2021science,shi2023surprising}. Our work can thus be seen as an early large-scale, empirical, data-driven study of \textit{interestingness} in the scientific domain, and as a step toward AI systems for scientific discovery that generate not only novel ideas, but ideas that scientists themselves find sufficiently interesting to warrant further thought and exploration.
\begin{figure*}[!t]
    \centering
    \includegraphics[width=1\linewidth]{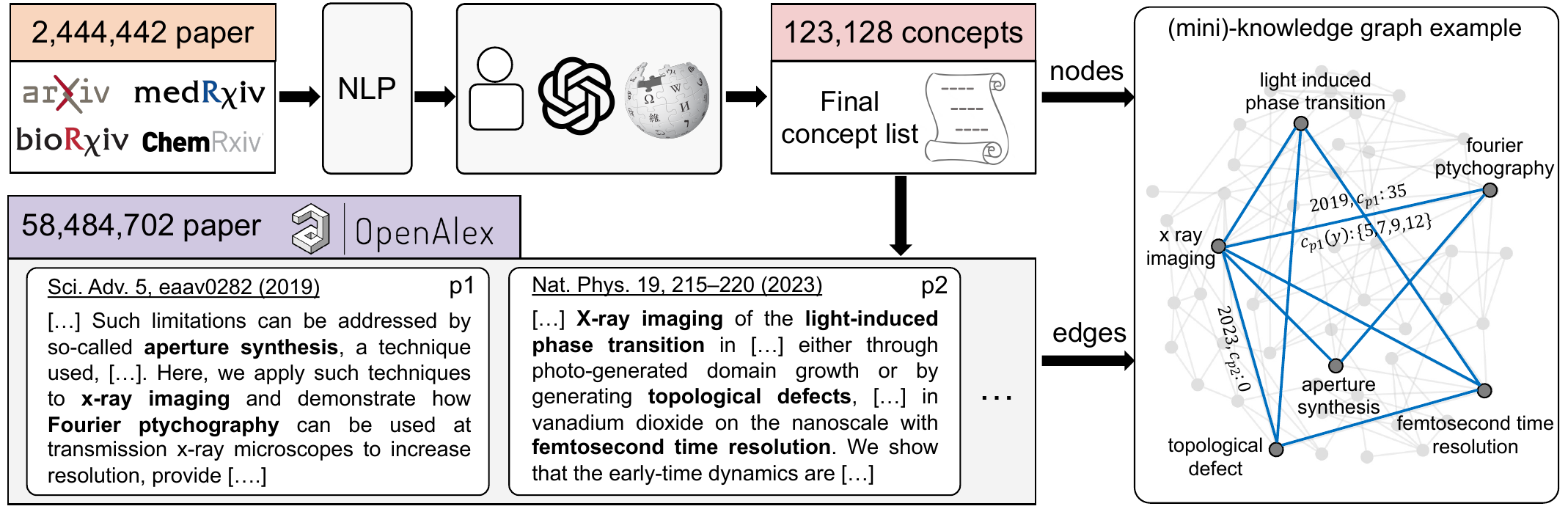}
    \caption{Knowledge graph generation. Nodes represent scientific concepts extracted from 2.44 million paper titles and abstracts using the Rapid Automatic Keyword Extraction (RAKE) algorithm \cite{rose2010automatic}, further refined with custom NLP techniques, manual review, GPT, and Wikipedia (to restore mistakenly removed concepts), resulting in a final list of 123,128 concepts. Edges are formed when two concepts co-occur in the paper (in title and abstract), across a corpus of more than 58 million papers from OpenAlex \cite{openalex}, augmented with citation data as a proxy for impact. A mini-knowledge graph (blue edges) is constructed from two randomly selected papers (p1 and p2) \cite{wakonig2019x, johnson2023ultrafast} from OpenAlex as an example. Here, $c_{p1}$ represents the total citations of paper p1 since its publication, and $c_{p1}(y)$ is its annual citations from 2019 to 2022 (e.g., $c_{p1}(2019)=12$). The citation value of an edge is the sum of the citation counts of all papers that create that edge.}
    \label{fig:kgraph}
\end{figure*}

\subsection*{Contribution 1: A Knowledge Graph Foundation for Generating Scientific Ideas}
To achieve fine-grained control over generating new research ideas --- for example, by tailoring suggestions to a scientist's specific interests or targeting properties such as predicted impact and novelty --- we ground our approach in a large knowledge graph constructed from scientific literature. This offers additional control compared to relying solely on publicly available large language models (such as GPT \cite{achiam2023gpt}, Gemini \cite{reid2024gemini}, or Claude \cite{anthropic2024claude3}), where the generation process is typically guided through prompt design. By building a personalized knowledge graph, we can systematically characterize a scientist's research interests. Importantly, the knowledge graph also enables us to analyze which factors correlate with expert interest and to predict it from graph features.

In Fig.~\ref{fig:kgraph}, we show the construction of the knowledge graph. The graph consists of vertices representing scientific concepts, with edges formed when concepts co-occur in paper titles or abstracts. These edges are weighted by citation and temporal information to capture scientific impact and trends over time. 

\textbf{Concept generation.} To construct the set of vertices, we analyzed approximately 2.44 million articles from arXiv, bioRxiv, ChemRxiv, and medRxiv, with a data cutoff in February 2023. The arXiv dataset is available on \href{https://www.kaggle.com/datasets/Cornell-University/arxiv}{Kaggle}, and metadata from the other preprint sources was obtained via their respective APIs. We processed article titles and abstracts using the Rapid Automatic Keyword Extraction (RAKE) algorithm \cite{rose2010automatic}, a statistical text analysis method for automatically identifying important keywords, enhanced with additional stopwords to extract candidate scientific concepts. Initial filtering retained two-word concepts (e.g., \textit{Gouy phase}) appearing in at least nine papers, and concepts with three or more words (e.g., \textit{recurrent neural network}) appearing in at least six papers, yielding an initial pool of 726,439 concepts. The thresholds were chosen to balance concept quality and coverage: lower thresholds introduce noise from RAKE artifacts, while higher thresholds exclude meaningful scientific terms. We applied a higher cutoff to two-word concepts because shorter phrases are more likely to produce noisy candidates than longer, more specific terms. The thresholds were not formally optimized, and alternative values may also work depending on the corpus and use case. A limitation of frequency-based filtering is that very recent or highly specialized concepts that appear rarely in the corpus may fall below the threshold and be excluded. How to better retain such emerging or niche concepts without increasing noise remains an interesting direction for future work.

To improve the quality of the concept list, we applied a set of rule-based filters to automatically remove obvious extraction errors introduced by RAKE. For example, we discarded entries containing words from predefined blocklists covering common verbs (e.g., `improve', `reduce'), adverbs (e.g., `significantly'), discourse markers (e.g., `however'), general nouns (e.g., `University'), month names, physical units, and special symbols. This automated filtering was followed by manual review by the authors, which removed remaining non-conceptual phrases, verb phrases, conjunctions, and proper nouns such as city and country names. This combined process reduced the set to 368,825 concepts (see Ref.~\cite{gu2025forecasting} for details). However, manually reviewing many entries was time-consuming, so we used GPT-3.5 to further refine the concept list through repeated filtering rounds. In each round, GPT-3.5 received a random batch of 10 concepts and instructed to retain only informative descriptors of research activity, discarding concepts that were too general, unclear, or grammatically incorrect, with a target removal rate of approximately one third per batch (see the Supporting Information for the full prompt). This process eliminated 286,311 entries. To recover valid concepts that may have been mistakenly excluded, we cross-referenced the removed entries against Wikipedia and restored 40,614 terms. The final list comprises 123,128 scientific concepts, which serves as the vertices of the knowledge graph. Our refinement process did not follow a formal optimization criterion; rather, it was intended to produce a concept set of sufficient quality while keeping the scale of graph construction manageable. \textcolor{black}{Future work could explore alternative concept-generation approaches, such as embedding-based clustering to group semantically similar phrases and reduce redundancy.}
\begin{figure*}[!t]
    \centering
    \includegraphics[width=1\linewidth]{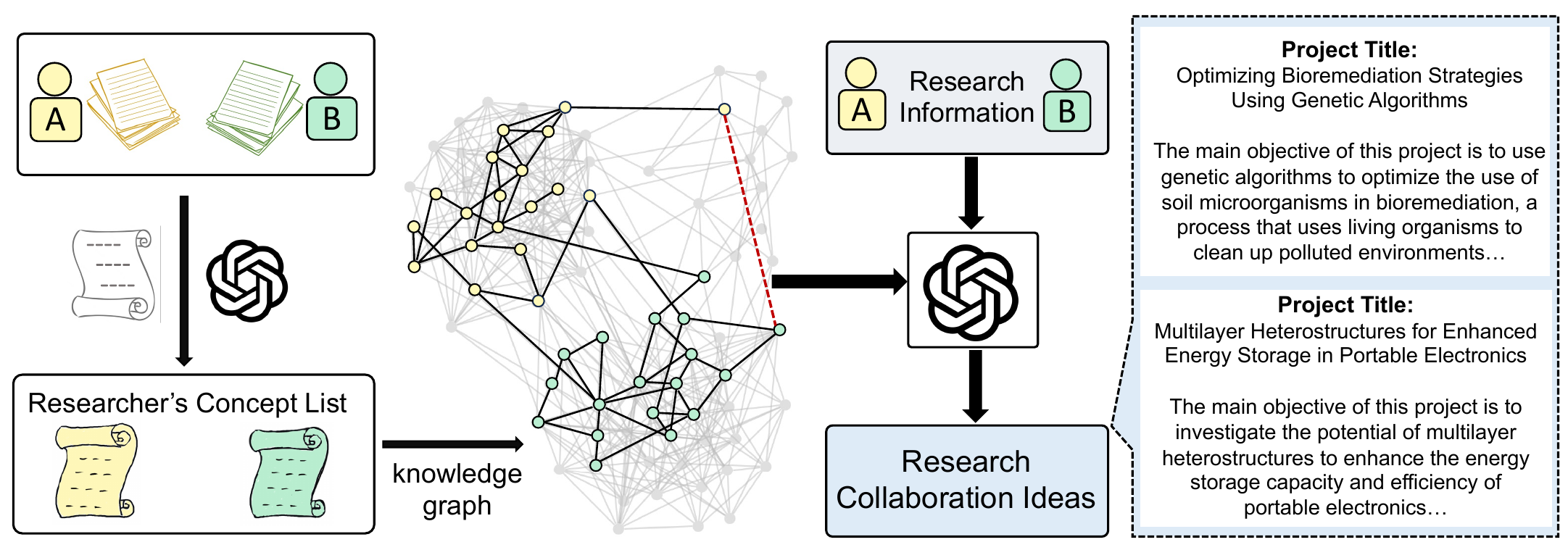}
    \caption{\textbf{\textsc{SciMuse} suggests research ideas or collaborations using a knowledge graph and an LLM}. We first extract concepts from the recent publications of Researchers A and B, refine them using GPT-4, and identify relevant sub-networks in the knowledge graph. We then use these concept pairs, together with information about the researchers' work, as input to GPT-4 to generate personalized research ideas or collaboration projects.}
    \label{fig:suggestion}
\end{figure*}

\textbf{Edge generation.} To construct edges augmented with citation data as a proxy for scientific impact, we used the open-source OpenAlex database (accessible via the \href{https://openalex.s3.amazonaws.com/browse.html}{OpenAlex S3 bucket}; also see \cite{openalex} for detailed documentation). We filtered the dataset to include only journal articles with available titles, abstracts, and citation data, resulting in approximately 92 million scientific papers. Of these, more than 58 million papers contained at least two concepts from the final concept list, thereby enabling edge construction in the knowledge graph. These edges contain information about the co-occurrence of concepts in scientific papers (titles and abstracts) and their citation rates. This graph representation was recently introduced in \cite{gu2025forecasting} to predict the impact of future research topics. Our final knowledge graph has 123,128 vertices and more than 168 million edges (built from \textcolor{black}{1.52 billion} edges, containing multi-edges when multiple papers create the same edge), with a data cutoff at April 2023. It captures part of the evolution of science from 1666 (a text by Robert Hooke on recent observations of Jupiter and Saturn \cite{hooke1some}, creating the first edge in the knowledge graph, connecting \textit{naked eye} and \textit{apparent diameter}) to April 2023. The detailed knowledge graph construction is shown in Fig.~\ref{fig:kgraph}. \textcolor{black}{We note that, in the knowledge graph, citation intent is not distinguished, and supporting, contrasting, and critical citations are therefore treated equivalently. Incorporating citation context might be an interesting future work.}

\subsection*{Contribution 2: Generating Personalized Research Directions Using a Knowledge Graph and LLMs}\label{se:idea_generation}

With the knowledge graph in place, we now describe how it is used to generate personalized proposals for collaborations between two scientists. To reflect each researcher’s current scientific interests, we prioritize their recent publications. Specifically, we collect titles and abstracts from papers for each researcher published in the past two years and include up to 20 papers, ordered from most to least recent. If fewer than three publications are available within this period, we expand the time window backward by one year at a time until at least three publications are obtained. These publications form the basis for constructing a concept-based profile for each scientist and generating personalized suggestions. We then extract candidate concepts from the titles and abstracts of these publications by matching them against the final concept list shown in Fig.~\ref{fig:kgraph}. The extracted concepts are further refined by GPT-4 to improve their quality, retaining only those that are meaningful and relevant to the scientist’s research direction (see the prompt in the Supporting Information). This process builds a personalized subgraph in the knowledge graph that reflects each scientist’s research direction (see Fig.~\ref{fig:suggestion}).

We then used a single, fixed prompt template across all generation conditions (see the full template in the Supporting Information). The prompt explicitly framed the task as follows: “\textit{Two researchers A and B, with expertise in \texttt{concept1} and \texttt{concept2} respectively, are eager to collaborate on a novel interdisciplinary project that leverages their unique strengths and creates synergy between their fields.}” The prompt provided up to seven recent paper titles per researcher and asked the LLM to propose a “\textit{creative and surprising scientific project.}” Across all conditions, the prompt template was held constant; only the source of concepts varied, namely whether the LLM was given a 
\textcolor{black}{high-citation-predicted} concept pair, a random concept pair, or no externally supplied concept pair \textcolor{black}{(i.e., titles-only GPT baseline)}. Rather than relying on a one-shot completion, we instructed the model to follow a self-reflective critique-and-revision procedure inspired by iterative self-feedback and verbal reflection methods \cite{shinn2023reflexion,madaan2024self}: it first briefly explained the guiding concepts, then generated candidate scientific contexts in which they might be combined, criticized these contexts with respect to how naturally and meaningfully they merged the two areas, and summarized these reflections across repeated rounds before returning a final project title, objective, and research questions. This staged self-reflection was intended to reduce superficial concept combinations and encourage more coherent and non-obvious integrations of the two researchers’ backgrounds. More elaborate prompt variants tested in small pilots did not improve expert-rated interestingness. Note that the \textcolor{black}{titles-only GPT baseline} relied only on the paper titles of the two researchers and did not provide an explicit concept pair to the LLM, serving as a useful sanity check.

Among the three generation conditions, one used concept pairs predicted to have high impact. To identify these pairs, we adapted the machine-learning framework of \cite{gu2025forecasting} to our large knowledge graph. Specifically, impact prediction is formulated as a binary classification task: given a pair of concepts that is not yet connected in the knowledge graph, the model predicts whether it will exceed a citation threshold within 3 years, which we use as a proxy for high impact (here, at least 10 citations). For each concept pair, we extract 141 graph features that capture both network structure (e.g., node degree, PageRank, and cosine similarity) and citation-based information from the knowledge graph. These features are then used as input to a fully connected neural network with four hidden layers of 600 neurons each, trained on historically unconnected concept pairs to predict whether the pair will exceed the chosen citation threshold; see Ref.~\cite{gu2025forecasting} for a detailed description of the approach.

\subsection*{Contribution 3: A Large-Scale Expert Evaluation of AI-Generated Ideas}
Having described the idea-generation pipeline, we now turn to evaluating the AI-generated proposals. A central contribution of this work is a large-scale assessment of the scientific interestingness of AI-generated research ideas by experienced scientists. To this end, we conducted an expert evaluation involving research group leaders from the Max Planck Society, one of the world’s largest and most prestigious research organizations. As these group leaders routinely evaluate research proposals and shape scientific directions, they are well positioned to assess the quality of AI-generated ideas.

\textbf{Choice of evaluation strategy --} 
We generate AI-written research project ideas framed as potential collaborations between two group leaders within the Max Planck Society, with one serving as the evaluator. We vary the pairing strategy by selecting researcher pairs (i) from the same institute, (ii) from the same broad domain (either natural sciences or humanities/social sciences), or (iii) across domains. This design allows us to assess AI-generated ideas in both disciplinary and interdisciplinary settings. Framing each suggestion as a collaboration between an evaluating researcher (Researcher A) and a proposed collaborator (Researcher B) was a deliberate design choice. Rather than generating projects for a single researcher, which would more strongly constrain the search to that person's current agenda, each suggestion combines the research profiles of two individuals and may span institutes or broad domains. Although this does not eliminate the risk of local relevance bias, it reduces the extent to which personalization simply extrapolates a single researcher's existing trajectory. \textcolor{black}{Each suggestion was shown only to Researcher A, with Researcher B’s name, institute, and other identifying information anonymized.}

We chose to collect an interest rating from only the evaluating researcher A for several reasons. First, early-stage ideas can be strategically sensitive, and presenting the same potentially high-value idea to multiple groups may raise concerns about inadvertent disclosure, as well as priority and credit assignment. Second, our primary outcome is whether an expert finds a suggestion scientifically interesting, which is a necessary precursor to collaboration. By contrast, mutual willingness to collaborate depends on many additional factors, such as timing, resources, and interpersonal preferences, that are outside the scope of our system and would confound interpretation. Third, bilateral evaluation would substantially increase logistical complexity and participant burden. In particular, it would require both researchers to consent to shared personalized suggestions and provide ratings; incomplete participation within a pair would further reduce the number of ideas that could be evaluated.

Evaluating each idea by both involved researchers would therefore require a different study design, including explicit bilateral consent, and protocols for handling priority/credit, a substantially increased evaluation burden, and procedures for cases in which one party responds while the other does not. We view this as an important direction for future work and expect the methods and dataset here will provide a helpful foundation. 
\begin{table}[!t]
\centering
\caption{\textbf{Recruitment results}}
\label{table:recruitment}
\begin{tabular}{lrrrr}
\textbf{Outcome} & \textbf{Total} & \textbf{Nat} & \textbf{Soc} & \textbf{\% of invited} \\
\midrule
All                & 1064 & 983 & 81 & -\\
Invited              & 267 & 244 & 23 & 100\%\\
Participated         & 110 & 104 &  6 & 41.20\% \\
Did not participate  & 157 & 140 & 17 & 58.80\% \\
\midrule
\multicolumn{4}{l}{\textit{\small Reasons for non-participation:}} & \textbf{\% of invited}\\
\hspace{2mm}No reply                 & 138 & 128 & 10 & 51.69\% \\
\hspace{2mm}Lost contact             &  12 &   8 &  4 &  4.49\% \\
\hspace{2mm}No time                  &   6 &   3 &  3 &  2.25\% \\
\hspace{2mm}Ethical/personal concern &   1 &   1 &  0 &  0.37\% \\
\bottomrule
\multicolumn{5}{p{0.97\linewidth}}{\textit{Notes:} Invitations were sent to research group leaders from 69 Max Planck Institutes (59 \textit{nat}, 10 \textit{soc}). Nat and Soc denote natural science institutes and social sciences/humanities institutes, respectively. Lost contact refers to invitees who initially expressed interest but did not perform the evaluation.}
\end{tabular}
\end{table}
\begin{table}[!t]
\centering
\caption{\textbf{Participant bibliometrics}}
\label{table:participant}
\begin{tabular}{lrrrr}
\textbf{ } & \textbf{Mean} & \textbf{Median} & \textbf{Min} & \textbf{Max} \\
\midrule
Number of papers    & 59.7   & 36.0   & 9   & 402   \\
Number of citations & 3759.7 & 1630.0 & 20  & 85778 \\
\bottomrule
\end{tabular}
\end{table}
\begin{figure*}[!t]
    \centering
    \includegraphics[width=0.93\linewidth]{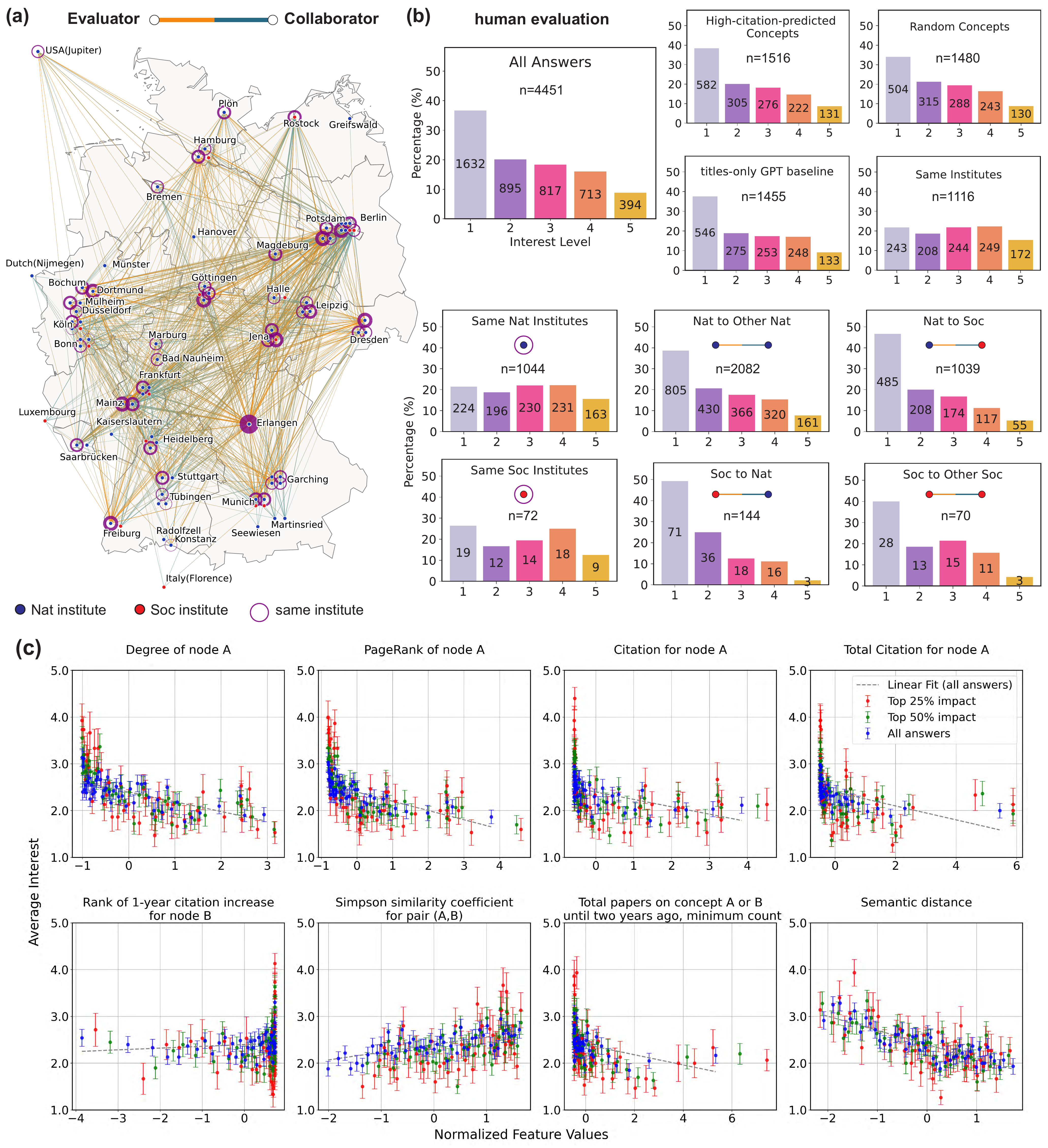}
    \caption{\textbf{Large-scale human evaluation within the Max Planck Society.} \textbf{(a)} \textcolor{black}{Locations of Max Planck Institutes and participating group leaders. The labels “Dutch (Nijmegen),” “Luxembourg,” “USA (Jupiter),” and “Italy (Florence)” denote institutes outside Germany rather than their map locations.} The Germany map is based on the Eurostat GISCO dataset \cite{eurostatNUTS}. A total of 4,451 AI-generated ideas were evaluated by 110 group leaders. Bicolored edges connect the evaluator (Researcher A, orange) and proposed collaborator (Researcher B, teal); edge thickness indicates the number of suggestions. Purple circles mark intra-institute collaborations, and blue and red dots natural-science (\textit{nat}) and social-science (\textit{soc}) institutes. Because of extensive edge overlap, panel (a) mainly shows the scale and geographic spread of the network. 
    \textbf{(b)}: Distribution of interest ratings on a 5-point scale from 1 (`not interesting') to 5 (`very interesting'). 
    In total, 394 suggestions received a rating of 5 and 713 a rating of 4. \textcolor{black}{Ratings are grouped by generation setting, institute, and research-domain pairing.} \textbf{(c)}: Correlations between eight knowledge-graph features and researchers' interest ratings \textcolor{black}{for 2,996 suggestions using the knowledge graph}. For each z-score-normalized feature, suggestions were ordered by feature value and divided into 50 equal-sized bins. The x- and y-axes show the mean feature value and interest score, with error bars indicating the standard error of the mean. Blue, green, and red points denote all 2,996 suggestions and the top 50\% and 25\% of concept pairs by predicted impact (adapted from \cite{gu2025forecasting}).}
    \label{fig:statistics}
\end{figure*} 

\textbf{Recruitment and participation --} We identified 1,064 group leaders with publicly available Google Scholar profiles (as of January 2024) across all 87 Max Planck Institutes. We then grouped the institutes into two broad categories: Class~1 (\textit{nat}), spanning natural sciences, technology, mathematics, and medicine (68 institutes), and Class~2 (\textit{soc}), spanning social sciences and humanities (19 institutes). This classification was performed manually based on institute titles and research focus, and cross-validated with GPT-4o, which produced the same assignments. From this pool, we randomly selected and invited 267 researchers (excluding directors) by email, across 69 institutes (59 \textit{nat}, 10 \textit{soc}).

Of the 267 invited group leaders, 110 participated (41.2\%), whereas 157 did not. Most non-participation resulted from non-response rather than explicit refusal: 138 invitees did not reply, 12 initially expressed interest but did not perform the survey, 6 declined due to lack of time, and 1 reported ethical concerns about AI-assisted idea generation. For those who agreed to participate, we prepared a personalized online survey containing their AI-generated suggestions and sent the survey link by email.  Participation was dominated by natural-science institutes, largely reflecting the composition of the invited pool, although the concentration was slightly higher among participants than among invitees (Table~\ref{table:recruitment}). We did not collect measures of prior attitudes toward AI at the time of invitation and therefore cannot directly assess whether participants and non-participants differed in their predispositions toward AI-generated research ideas.

Among the 110 participating group leaders, 104 are from natural science institutes and 6 from social science institutes, spanning a total of 54 Max Planck Institutes (Fig.\ref{fig:statistics}(a)). As of January 1, 2024, when the evaluations were conducted, \textcolor{black}{the typical participant had published 36 papers (median; mean 59.7, range 9–402) and accumulated 1,630 citations (median; mean 3,759.7, range 20–85,778). The higher means relative to the medians, together with the large upper ranges, indicate that both distributions are right-skewed (Table~\ref{table:participant}).}

\textbf{Expert evaluation results --} For each participant, we provided 48 personalized research project suggestions created under the three settings described in Contribution~2 (\textcolor{black}{high-citation-predicted} concept pair, random concept pair, or \textcolor{black}{titles-only GPT baseline}; 16 suggestions per setting). The researchers rated each AI-generated ideas on a 5-point scale ranging from 1 (`not interesting') to 5 (`very interesting'). The scale was chosen to capture the subjective interest evaluation of the expert evaluator. \textcolor{black}{Of the 5,280 possible ratings from 110 participants, 4,451 were recorded in the end, corresponding to an 84.3\% completion rate. In Fig.~\ref{fig:statistics}(b), the mean interest score was 2.403$\pm$0.020, below the midpoint of the 1–5 scale, the most common rating was 1, received by 1,632 ideas (36.7\%). A total of 1,107 ideas (24.9\%) received a high-interest rating of 4 or 5, including 394 rated 5.}

In Fig.~\ref{fig:statistics}(b), \textcolor{black}{we compare the interest ratings of AI-generated ideas across the three generation conditions described in Contribution~2.} The mean interest score was \(2.350 \pm 0.034\) for ideas based on \textcolor{black}{high-citation-predicted} concepts, \(2.446 \pm 0.035\) for ideas based on random concepts, and \(2.414 \pm 0.036\) for the \textcolor{black}{titles-only GPT baseline}, where the uncertainty denotes the standard error of the mean (SEM). \textcolor{black}{We find weak statistical evidence (1.94$\sigma$) that ideas generated from high-citation-predicted concept pairs are rated less interesting than those generated from random concept pairs. This points to a broader tension between predicted citation impact and empirical expert-rated interest. We therefore compared how both quantities depend on knowledge-graph features and found opposite trends for several prominent features. For example, higher citation counts of individual concepts are associated with higher predicted citation impact but lower empirical interest ratings. This suggests that predicted citation impact and expert-rated interest appear to capture distinct and, for some graph features, oppositely aligned properties of scientific ideas. See Supporting Information for details.}
Likewise, the fraction of ideas receiving a high-interest rating (4 or 5 out of 5) was comparable across conditions: \((23.3 \pm 1.09)\%\) for ideas based on \textcolor{black}{high-citation-predicted} concepts, \((25.2 \pm 1.13)\%\) for ideas based on random concepts, and \((26.2 \pm 1.15)\%\) for the \textcolor{black}{titles-only GPT baseline}. 
\begin{figure}[!t]
    \centering
    \includegraphics[width=1\linewidth]{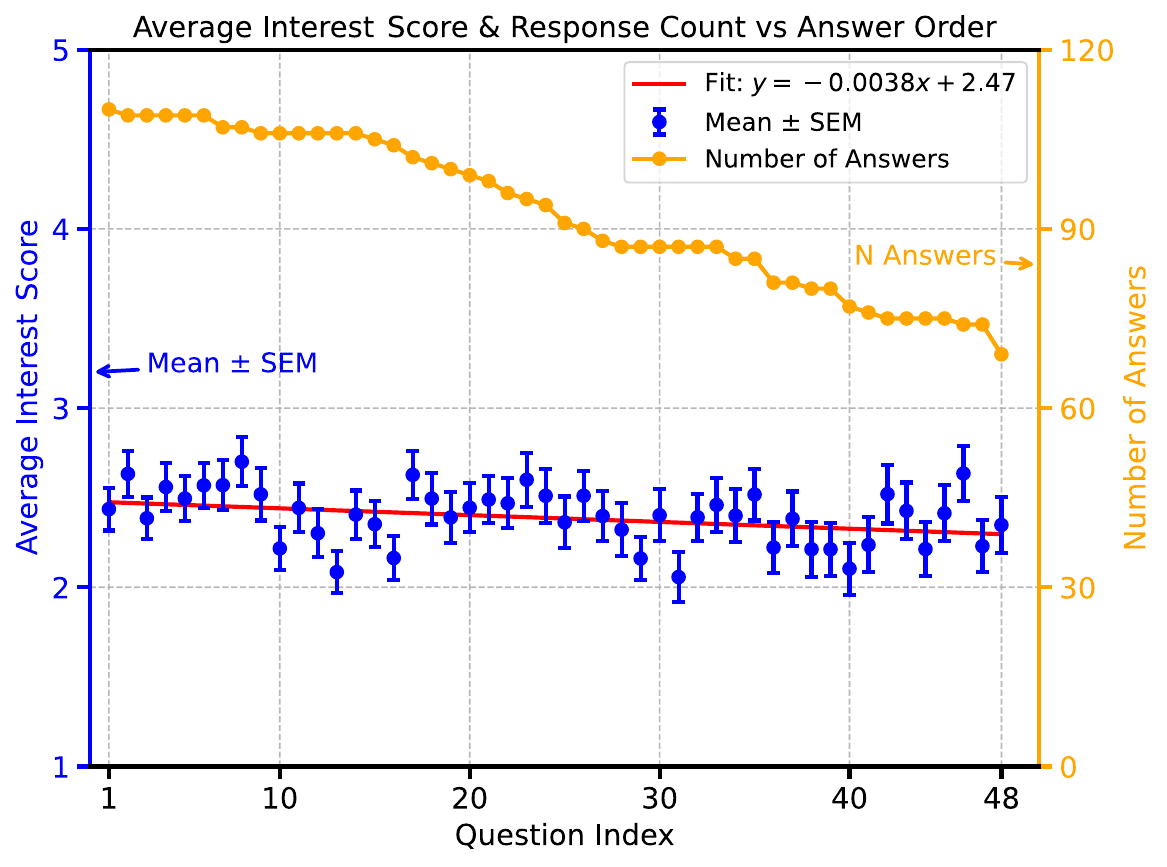}
    \caption{\textbf{Mean Interest Score and Response Count vs. Question Order} The number of recorded responses decreases with question index (orange; right axis), indicating that some participants did not complete all 48 proposals \textcolor{black}{and resulting in an overall completion rate of 84.3\%.} For each question position, we plot the mean interest rating across all available responses (blue; left axis). SEM is the standard error of the mean. A linear fit to the mean ratings (red) gives $y=kx+d$ with $k=-0.0038\pm0.0015$ and $d=2.475\pm0.037$ ($R^2=0.1124$), corresponding to a small change of approximately 0.18 rating points across 48 questions. Overall, this indicates only a weak order/fatigue effect relative to the 1--5 rating scale.}
    \label{fig:answer_order}
\end{figure}

Ratings could vary with question order, potentially biasing aggregate estimates of interest. To see whether the evaluation procedure introduced such bias, we examined order effects in the questionnaire. Specifically, we analyzed ratings across the questionnaire, tracking both the mean score (with standard error of the mean) and the number of responses for each question. The number of responses decreases for later questions, whereas the mean rating remains fairly consistent throughout the questionnaire (see Fig.~\ref{fig:answer_order}). This suggests that question order did not strongly influence participants' ratings. 

\textcolor{black}{Taken together, these results provide no evidence that supplying either random or high-citation-predicted concept pairs improves expert-rated interest over the \textcolor{black}{titles-only GPT baseline}. However, the structured graph representation provides explicit control over which scientific concepts guide generation and contains predictive information about expert-rated interest in Contribution~4. This suggests a future direction in which concept pairs are selected based on predicted interest rather than predicted citation impact. Whether such interest-optimized concept selection can prospectively increase expert-rated interest remains an open question.}

\textbf{Graph features and interest ratings --} 
We analyzed 2,996 research suggestions using \textcolor{black}{high-citation-predicted and random} concept pairs from the knowledge graph. \textcolor{black}{The titles-only GPT baseline was excluded because these suggestions were not associated with concept pairs and therefore lacked the corresponding knowledge-graph features.} Suggestions were sorted by each graph feature and divided into 50 equal-sized bins, for which mean interest scores and standard errors were calculated. Fig.~\ref{fig:statistics}(c) shows the resulting correlations between graph features and interest ratings, based on linear fits to the binned means. For example, the vertex degree and PageRank of the first concept from the evaluating researcher's concept list both show negative slopes with interest: \(k=-0.213\pm0.030\) (\textcolor{black}{binned} \(R^2=0.480\), \textcolor{black}{idea-level Spearman correlation \(\rho=-0.166\)}) for degree and \(k=-0.192\pm0.028\) (\(R^2=0.474\), \textcolor{black}{\(\rho=-0.176\)}) for PageRank. This suggests that the more connected a concept is within the knowledge graph, the less interesting the resulting research project tends to be. A similar behaviour appears for citation-based quantities, such as citations in the last year (\(k=-0.162\pm0.034\), \textcolor{black}{\(R^2=0.320\), $\rho=-0.143$}) and total citations (\(k=-0.143\pm0.028\), \textcolor{black}{\(R^2=0.330\), $\rho=-0.153$}). In addition, the semantic distance between researchers also shows a negative relationship with interest (\(k=-0.309\pm0.027\), \textcolor{black}{\(R^2=0.734\), $\rho=-0.223$}).

\textcolor{black}{Semantic distance between the evaluator and proposed collaborator shows a negative correlation with the evaluator's interest rating. As the evaluator is one of the two researchers defining this distance, semantic distance is inherently coupled to self-relevance: more distant collaborations are also further from the evaluator's own research area. The present design therefore cannot distinguish an effect of interdisciplinarity itself from a preference for ideas closer to the evaluator's expertise. The self-relevance mechanism may also contribute to higher ratings for some more closely related researcher pairings in Fig. 4(b).}
We show these correlations for all 2,996 ratings, as well as for the top 50\% and top 25\% of high-citation-predicted concept pairs in Fig.~\ref{fig:statistics}(c). This suggests that some correlations become more pronounced when ideas are generated using \textcolor{black}{high-citation-predicted} concept pairs.

\begin{figure}[!t]
    \centering
    \includegraphics[width=1\linewidth]{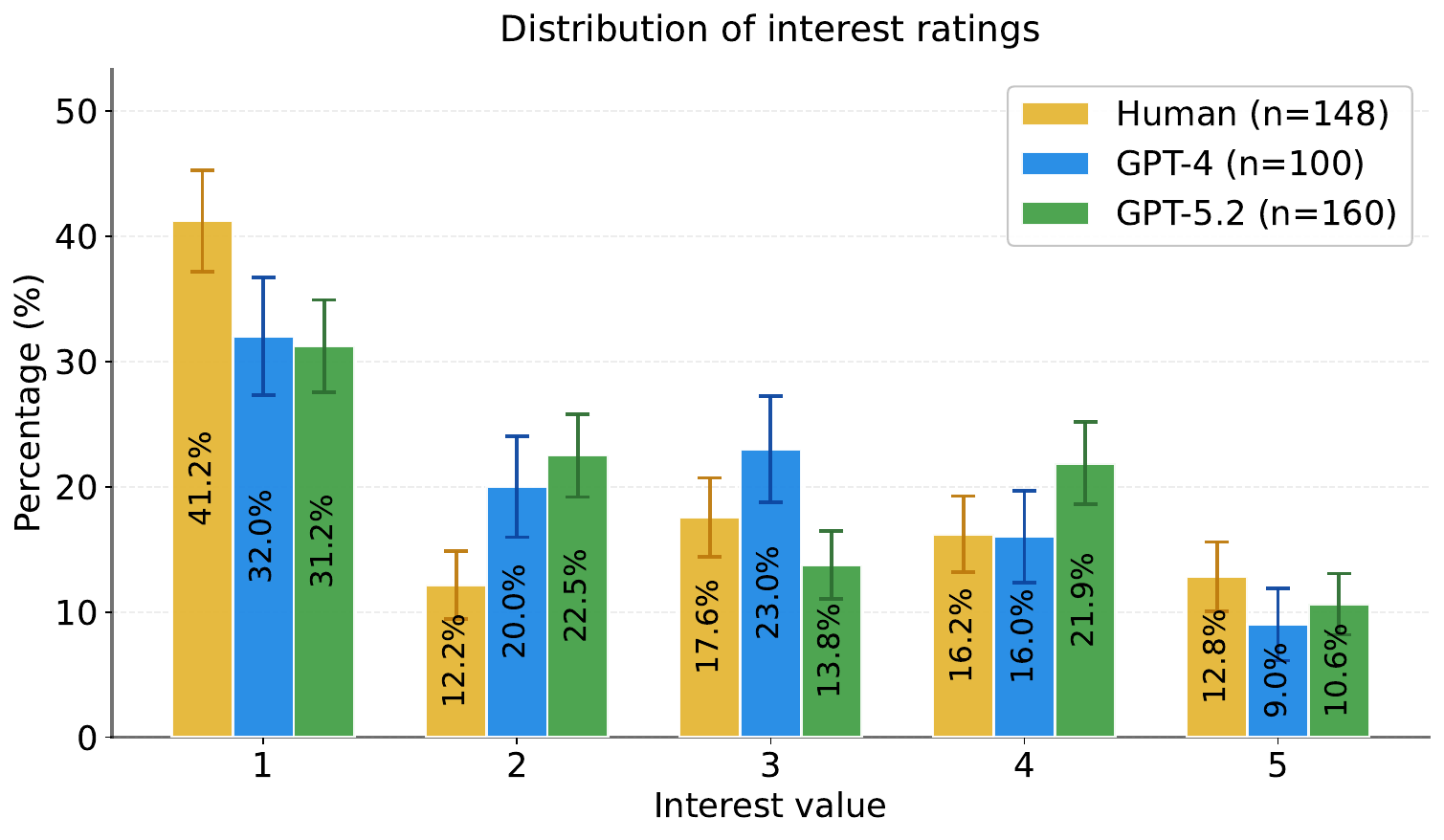}
    \caption{\textbf{Human-generated and GPT-generated ideas.} Distribution of interest ratings on a 5-point scale from 1 (`not interesting') to 5 (`very interesting') for human-generated ideas, GPT-4-generated ideas, and GPT-5.2-generated ideas. The human-written baseline consists of 148 personalized ideas written by seven \textcolor{black}{experienced PhD students} for six research group leaders from physics-related fields. GPT-5.2 contributed 160 additional personalized ideas, and GPT-4 ideas (100 in total) from the main evaluation were restricted to the same six evaluators and assessed under the same protocol. Bar heights show the percentage of ideas receiving each rating, with percentages indicated inside the bars. The distributions are similar across all three sources, with comparable fractions of highly rated ideas (ratings of 4 or 5): \((29.1 \pm 3.7)\%\) for human-generated ideas, \((25.0 \pm 4.4)\%\) for GPT-4, and \((32.5 \pm 3.7)\%\) for GPT-5.2. Error bars denote the standard error of the proportion.}
    \label{fig:humanGenIdeas}
\end{figure}
\textbf{Baseline of human-generated ideas --} To assess how AI-generated proposals compare with human-written ones under \textcolor{black}{matched prompt-input conditions,} 
we constructed an additional \textcolor{black}{small-scale baseline of human-written research ideas, which we consider as an useful sanity-check}. Specifically, we asked seven \textcolor{black}{experienced PhD students} (three women and four men, all in at least the third year of their PhD) to write personalized research ideas for a subset of six research group leaders \textcolor{black}{from physics-related fields, limiting the broader representativeness of this baseline}. The human idea generators worked in scientific domains related to those of the evaluators, but did not know their identities. For each task, they received the same type of input information as the LLM-based system: a set of articles by two researchers (one of them being the target evaluator PI), together with two scientific concepts and their GPT-generated definitions. They were instructed to generate an interesting research idea and were given 10 minutes per task without access to the internet, phones, computers, or any other external resources. Each PhD student was assigned up to 30 tasks, each requiring them to produce a project title and a short description including the underlying research question. In total, this yielded 148 human-generated research ideas. In addition, we further generated 160 research ideas using the new model GPT-5.2 with the same input information. \textcolor{black}{Note that, under matched prompt-input conditions, humans were limited to the provided materials and their own knowledge, whereas LLMs could draw on broader pretraining knowledge. This asymmetry potentially favors the AI systems and does not constitute a fully matched head-to-head comparison.}

These human-generated and GPT-5.2-generated ideas were then evaluated by six research group leaders, all working in areas related to physics. The evaluators were not informed of the source of any individual idea. The resulting distribution of interest ratings is shown in Fig.~\ref{fig:humanGenIdeas}. For comparison, we also included the corresponding GPT-4-generated ideas from the evaluation reported in Contribution~3, restricted to the same six evaluators and assessed under the same protocol (100 ideas in total). The mean interest score was \(2.47 \pm 0.12\) for human-generated ideas, \(2.50 \pm 0.13\) for GPT-4-generated ideas, and \(2.58 \pm 0.11\) for GPT-5.2-generated ideas, where the uncertainty denotes SEM. Overall, the rating distributions were similar across the three sources. Likewise, the fraction of ideas receiving a high-interest rating (4 or 5 out of 5) was comparable across conditions: \((29.1 \pm 3.7)\%\) for human-generated ideas, \((25.0 \pm 4.4)\%\) for GPT-4-generated ideas, and \((32.5 \pm 3.7)\%\) for GPT-5.2-generated ideas. \textcolor{black}{These results cannot distinguish the quality of human and AI-generated ideas at a statistically significant level, but should be considered as a useful sanity-check.}

\begin{figure*}[!t]
    \centering
    \includegraphics[width=1\linewidth]{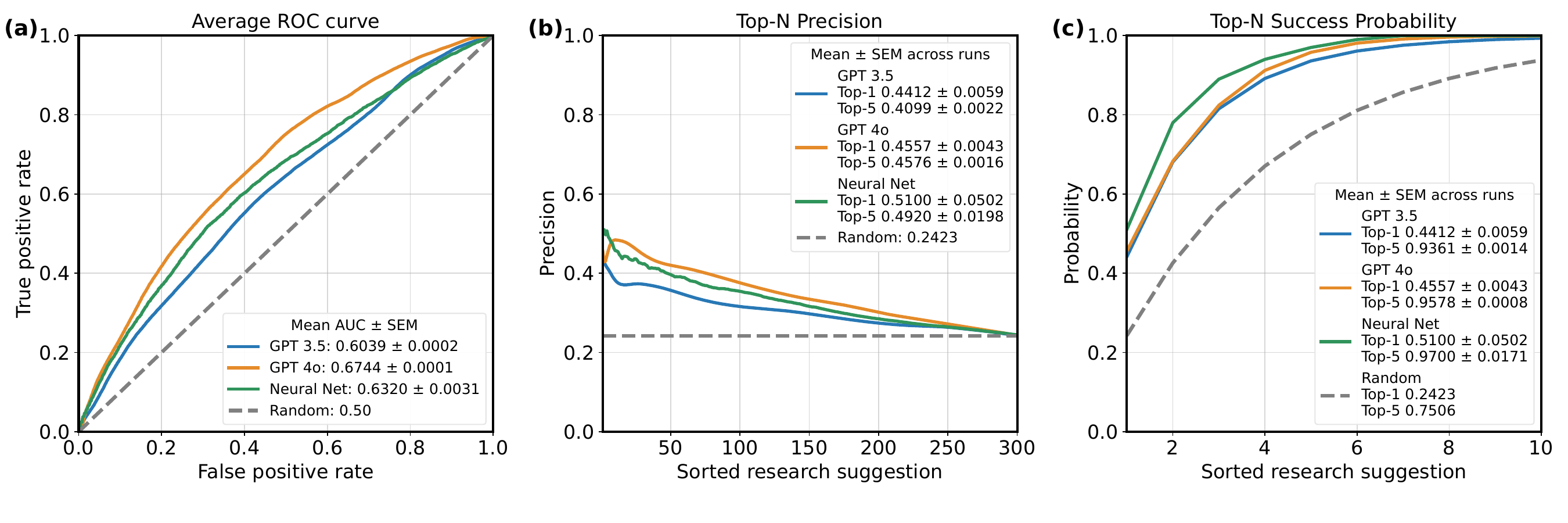}
    \caption{\textbf{Predicting scientific interest.} We use two methods to predict high-interest suggestions (ratings of 4 or 5): (1) a supervised neural network trained on human evaluations using 25 knowledge-graph features, without access to suggestion text, and (2) zero-shot GPT ranking based on pairwise comparisons with personalized research details. The neural network is evaluated using Monte Carlo cross-validation, while GPT rankings are obtained through an ELO-based tournament. \textbf{(a)}, The ROC curve gives an \textcolor{black}{AUC of \((63.20 \pm 0.31)\%\) for the neural network, \((67.44 \pm 0.01)\%\) for GPT-4o, and \((60.39 \pm 0.02)\%\) for GPT-3.5. \textbf{(b)}, Precision for the top-N suggestions is higher than random selection, with top-5 precision of \((49.20 \pm 1.98)\%\) for the neural network, \((45.76 \pm 0.16)\%\) for GPT-4o, and \((40.99 \pm 0.22)\%\) for GPT-3.5, and top-1 precision of \((51.00 \pm 5.02)\%\), \((45.57 \pm 0.43)\%\) and \((44.12 \pm 0.59)\%\) respectively.} \textbf{(c)}, The probability of having at least one high-interest suggestion among the top N recommendations is higher for the supervised neural network than for random selection. In practice, evaluation data from experienced researchers may not always be available; it is therefore encouraging that LLMs can still rank suggestions effectively, ranking the most interesting ideas first, even without human evaluation.}
    \label{fig:learning}
\end{figure*}
\subsection*{Contribution 4: Machine Learning Approaches to Predicting Scientific Interest}
The large-scale expert evaluation not only assessed the quality of AI-generated ideas, but also revealed that knowledge-graph features correlate with expert interest ratings. \textcolor{black}{This can be seen by the correlation of interest-evaluations with features of the underlying knowledge graph, in Fig.~\ref{fig:statistics}(c). For example, the semantic distance is negatively correlated and the Simpson similarity is positively correlated with interest-values. This naturally raises the question of whether high-interest research ideas (ratings of 4 or 5 out of 5) could be predicted, as these correlations could be learned by machine learning models.} To address this, we applied two methods to the 2,996 suggestions generated using concept pairs from the knowledge graph. First, we trained a neural network on researchers' responses using knowledge graph properties and human-evaluation rankings, without incorporating the GPT-generated text. Second, we used GPT in a zero-shot manner to rank the 2,996 suggestions independently of human evaluations. Both methods predicted 
interest despite relying on only partial information
(see Fig.~\ref{fig:learning}).  
\textcolor{black}{This shows that graph features contain predictive information about expert-rated interest and could therefore be used to rank concept pairs according to predicted interest. Whether such selection prospectively improves generated ideas remains to be tested.} Likewise, GPT-based zero-shot ranking is especially valuable when human evaluation data are unavailable.

For the supervised neural network, we used knowledge graph features to predict whether a research proposal would receive a high rating (4 or 5) or a lower rating (below 4). Given the limited amount of training data (2,996 data points, each corresponding to one research group leader's evaluation of a proposal's interest level), we adopted a low-data machine-learning approach using a small neural network with 25 selected high-performing input features, 50 neurons in a single hidden layer, and one output neuron. Dropout was used during training \cite{srivastava2014dropout}.\textcolor{black}{We used Monte Carlo cross-validation \cite{xu2001monte} (also known as repeated random sub-sampling validation) with 100 random splits. In each split, 2,247 (75\%) examples were used for training, 449 (15\%) for validation, and 300 (10\%) for testing. Feature normalization and feature selection were performed independently within each split using the training data only. The 144 candidate knowledge-graph features were ranked according to their individual predictive performance on the training set, and the 25 highest-ranking features were retained for model training.} Decision trees \cite{breiman2017classification} were also tested but did not outperform the neural network approach (details see the Supporting Information).

In the second method, we asked GPT-3.5 and GPT-4o to rank 2,996 suggestions from most to least interesting. Because the suggestions are personalized, we included research details, such as recent paper titles, in each prompt. Specifically, GPT was asked to compare pairs of randomly selected suggestions and determine which would be more interesting for the evaluator, with ties disallowed (forced choice). This comparison was repeated between 22,000 and 45,000 times, depending on the GPT version, and the suggestions were ranked using an ELO system, with each suggestion initialized at a score of 1400. The model's choices updated these rankings, resulting in a sorted list based on predicted interest.

For the binary classification task -- classifying research ideas as highly interesting (4 or 5 out of 5) or low-interest (3 or below) -- we used the Area Under the Curve (AUC) of the receiver operating characteristic (ROC) curve \cite{fawcett2004roc}. \textcolor{black}{The average AUC was \((67.44\pm 0.01)\)\% for GPT-4o, \((60.39\pm 0.02)\)\% for GPT-3.5, and \((63.20\pm 0.31)\)\% for the supervised neural-network model (Fig.~\ref{fig:learning}(a)).} 
More importantly, precision at the top of the ranking is especially relevant for our task, since we want to suggest highly interesting projects within a small subset of all ideas. We thus calculated precision for the top-\(N\) predicted suggestions. For small \(N\) (for example, \(N=5\)), the supervised approach achieved \((49.20 \pm 1.98)\%\), while GPT-4o and GPT-3.5 reached \((45.76 \pm 0.16)\%\) and \((40.99 \pm 0.22)\%\), respectively. This means that 49.2\% of the top-5 suggestions selected by the supervised model were rated as highly interesting, which is higher than random selection, as shown in Fig.~\ref{fig:learning}(b). We also measured the probability of finding at least one highly interesting suggestion within the top-\(N\) suggestions. In Fig.~\ref{fig:learning}(c), the supervised method yields a higher probability of identifying interesting ideas among the first few suggestions than a random approach. Moreover, GPT -- even without access to human evaluation data -- also ranks interesting suggestions higher than the random way. This capability is valuable in scenarios where human evaluations are costly or unavailable. It is also encouraging that the newer and more capable model GPT-4o performs better at predicting human interest than the earlier GPT-3.5.

\textcolor{black}{Overall, despite the target here being a very noisy and evaluator-subjective quantity, we see that both supervised and zero-shot methods can extract a weak, but non-negligible signal. This suggests that future, more powerful machine learning models trained on more data could predict scientific interest and thereby hopefully select autonomously interesting future research ideas.}

\subsection*{Discussion and Limitations}
We demonstrate how to generate personalized research ideas and evaluate their scientific interest through a large-scale expert evaluation of AI-generated research ideas, involving more than 100 research group leaders across diverse scientific domains, including the natural sciences and the humanities. The resulting dataset, published with this article, offers a valuable resource for the AI community. It contains thousands of AI-generated ideas, each annotated with an interest rating from highly qualified human expert at the level of research group leader or professor, and thus provides a rich foundation for developing and refining systems that can generate truly interesting scientific ideas.

Using this dataset, we develop two distinct methods for predicting the interest level of AI-generated ideas. First, we train a supervised neural network that predicts interest ratings using knowledge graph data alone, without access to the full suggestion text. This approach enables us \textcolor{black}{to rank abstract topics by predicted expert interest.} Second, we show that LLMs can autonomously rank the interest of ideas with good quality, even without human evaluation data. This capability is valuable when human evaluations are costly or unavailable. Furthermore, we observe that ranking quality improves with more advanced LLMs, which is encouraging for future developments. As publicly available models such as GPT \cite{achiam2023gpt}, Gemini \cite{reid2024gemini}, LLaMA \cite{meta2024llama3}, and Claude \cite{anthropic2024claude3} continue to improve rapidly \cite{chiang2024chatbot} -- especially in scientific domain knowledge \cite{wellawatte2023extracting, liang2024can} -- we anticipate that personalized research ideas may become increasingly targeted and relevant.

\textbf{Potential data-source bias --} It is important to note that the underlying knowledge graph is constructed from arXiv, bioRxiv, medRxiv, chemRxiv, and OpenAlex. This selection may introduce biases toward certain scientific disciplines and predominantly English-language literature. We mitigated data-level biases through a multi-stage concept-refinement process and a combination of automated and manual validation steps. While these reduce noise and obvious imbalances, residual biases inherent in both the publication databases and LLM-based text generation could not be eliminated completely.

\textbf{Potential institutional bias (Max Planck Society) --} Although our evaluation involved highly qualified researchers from a broad range of fields, all participants were affiliated with the Max Planck Society, and the sample was strongly dominated by the natural sciences. This matters because the Max Planck Society reflects a particular research culture and funding environment that may shape how researchers perceive speculative, interdisciplinary, or AI-generated ideas. Researchers in other settings -- for example, universities with different resource constraints, industrial laboratories, or institutions in other countries -- may apply different standards when judging whether an idea is interesting or worth pursuing. Our results should therefore not be interpreted as fully representative of the global scientific community, but rather as evidence from a large expert sample within a specific and highly research-intensive institutional context. Further studies across other institutions, countries, and disciplinary communities will be necessary to assess the broader generalizability of these findings.

\textbf{Potential self-selection bias among evaluators --} Because participation in the evaluation was voluntary, we cannot exclude the possibility of self-selection bias. Researchers who chose to participate may have been more curious about AI-assisted idea generation, more open to speculative or interdisciplinary suggestions, or more willing to engage in a time-intensive evaluation task than those who did not participate. This could affect the absolute level of interest ratings and limit the extent to which the results generalize beyond the participating sample. At the same time, most non-participation was due to non-response or practical constraints rather than explicit objections, so the direction and magnitude of this potential bias cannot be determined from our data.

\textbf{Subjectivity of interest ratings --} Scientific interest is inherently evaluator-dependent, and our 1--5 scale was designed to capture this personalized signal rather than a universal measure of idea quality. Different researchers likely used the scale with different thresholds, which limits the direct comparability of absolute scores across evaluators. However, because each evaluator rated multiple proposals, the ratings still provide a meaningful within-evaluator ordering of ideas from less to more interesting. The fact that these judgments can be predicted above chance (see Fig.~\ref{fig:learning}) suggests that they reflect systematic preferences rather than arbitrary noise.

\textcolor{black}{\textbf{Semantic distance and self-relevance --}
A limitation of the analysis is that, because each evaluator was one of the two researchers defining the proposed collaboration, the semantic distance to the collaborator is necessarily also a measure of distance from the evaluator's own research area. We thus cannot determine whether the observed negative correlation reflects an intrinsic effect of interdisciplinary distance, greater familiarity or relevance to the evaluator, or both. Establishing an evaluator-independent effect of field distance would require a different design in which ideas are rated by researchers who are not parties to the proposed collaboration.}

\subsection*{Outlook}
The methodologies employed by \textsc{SciMuse} have the potential to inspire novel and unexpected cross-disciplinary research on a broader scale. By analyzing millions of scientific papers across diverse disciplines, \textsc{SciMuse} may facilitate the discovery of potentially interesting collaborations between scientists from different fields that might otherwise remain unnoticed. This capability is particularly valuable in light of empirical evidence suggesting that interdisciplinary research could lead to impactful and award-winning outcomes \cite{uzzi2013atypical, rzhetsky2015choosing, fortunato2018science, wang2021science}. Accordingly, large scientific organizations, national funding agencies, and other stakeholders may benefit from adopting methodologies in the spirit of \textsc{SciMuse} to support highly interdisciplinary collaborations and research ideas that might otherwise remain untapped. In this way, such approaches have the potential to advance the progress and impact of science at a larger scale.

\textbf{Impact beyond citations --} In the present study, we used expected future citation counts as a scalable proxy for one possible downstream dimension of scientific impact. However, citations are field-dependent, sensitive to social and institutional dynamics, and do not fully capture other dimensions of scientific value, such as originality, disruptive potential, or relevance for technology, policy, and clinical practice \cite{barabasi2018formula,fortunato2018science,aksnes2019citations,aksnes2023citation,milojevic2025science}. Future work should therefore move toward a broader conception of impact by complementing citation-based forecasts with alternative targets, such as indicators of novelty or disruption, links to patents and technological applications, and references in policy documents or clinical guidelines \cite{bornmann2012measuring,wu2019large,wang2021science,lin2023sciscinet}.

\textbf{Trend prediction beyond established concept pairs --} An important next step is to move beyond representations based solely on a list of established concepts (in our case, concepts that occur nine or more times) and pairwise concept links. Such approaches naturally favor topics and relations that are already repeatedly mentioned in the literature and may therefore miss promising concepts that are only beginning to emerge \cite{salatino2017topics,salatino2018augur}. More generally, scientific opportunities are not always well represented by a single concept pair, but may instead involve higher-order combinations or hierarchical relations between topics \cite{shi2015weaving,benson2018simplicial,lin2023sciscinet}. This suggests that future systems should integrate pairwise link prediction with early topic-emergence detection, higher-order modeling, and ontology-based representations \cite{salatino2019cso,peng2023knowledge} to better capture surprising, impactful, and interesting research directions.

\textbf{Generating and evaluating collaborative suggestions --} Although we evaluated ideas from the perspective of an individual researcher, the same framework can be adapted to collaboration-oriented idea generation \cite{milojevic2014principles}. This introduces two additional design choices: (i) how to select pairs of researchers (e.g., based on overlap or complementarity of expertise), and (ii) how to choose the specific concepts or topics that guide idea generation \cite{guimera2005team, zeng2021fresh}. In the present work, we randomly selected pairs of researchers within the Max Planck Society using three pairing classes (same institute, same broad domain, or across domains), and formed concept pairs by taking one concept from each researcher and ranking the resulting pairs by expected impact. In a collaboration-focused setting, both steps could be made more targeted: researcher pairs could be selected deliberately, and the guiding topics could be chosen to define a clearly scoped shared direction (e.g., a particular method, dataset, or phenomenon). Generating ideas around carefully chosen topics should make the resulting suggestions more likely to be relevant to both researchers. The evaluation and interest-prediction methods introduced in this work could then be used to prioritize the most promising candidates before sharing them with both parties. Evaluating collaboration suggestions in such a two-sided setting would require explicit consent from both participants, since it involves sharing personalized and potentially sensitive suggestions.

\textbf{Generating ideas for individual researchers --} So far, we have generated ideas for pairs of researchers, implicitly assuming that the distance between their research fields introduces diversity into the generated ideas. One could also generate ideas for individual researchers \cite{sourati2023accelerating} and introduce diversity more explicitly, for example by identifying structural holes in their research focus (e.g., a technique that would meaningfully complement their work \cite{hargadon2002brokering, burt2004structural}), which could then be developed into full-fledged research proposals.

\textbf{Diversifying the scientific idea space --} A key challenge for AI systems that generate research ideas is to avoid the narrowing of the scientific search space \cite{lee2024empirical,doshi2024generative,hao2026artificial}. If ideas are optimized only for the current interests of a single researcher, the system may produce suggestions that remain close to that person’s existing agenda. In the present work, we mitigated this risk of narrowing the idea space by generating ideas for pairs of researchers (potentially from semantically distant fields). An important question for future AI ideation systems is how to generate research ideas for individual researchers while maintaining high personalized interest and increasing the diversity of the generated suggestions.

\textbf{Responsible autonomous scientific ideation --} As AI systems move toward more autonomous forms of scientific ideation and evaluation, ethical and institutional questions become increasingly central: who receives credit for AI-assisted ideas, how priority and intellectual property should be handled \cite{ouellette2025will}, and how such systems should be disclosed and governed \cite{binz2025should}. It will be important to ensure that automated proposal generation expands rather than overwhelms or homogenizes the scientific landscape \cite{he2026academic}, and that it leads to understanding rather than mere prediction \cite{krenn2022scientific,messeri2024artificial,barman2024towards}, so that increasingly autonomous AI broadens rather than diminishes the diversity and integrity of science \cite{blau2024protecting,tang2025risks}.

\textbf{Connections to simulators and automated experiments --} A particularly exciting long-term direction is to augment AI-based ideation with the automated implementation and evaluation of ideas. This would require connecting AI ideation systems to scientific tools such as physical simulators and design frameworks, including QuTiP \cite{lambert2026qutip}, Qiskit \cite{javadi2024quantum}, Tequila \cite{kottmann2021tequila}, PyTheus \cite{ruiz2023digital}, and Meta-Design \cite{arlt2026meta}, or to automated experimental platforms \cite{abolhasani2023rise,boiko2023autonomous,dai2024autonomous,m2024augmenting,leong2025steering}. In most current systems, the main experimental idea is still provided by human scientists. Looking further ahead, one could envision a more grounded and closed-loop form of scientific discovery, in which AI systems generate an interesting idea, connect it to simulators or experimental platforms, and autonomously test, refine, and implement it in physical systems. A few narrow prototypes point in this direction \cite{villaescusa2025denario,mitchener2025kosmos,arlt2025towards}. Such systems could have great long-term potential, as they could help close the scientific discovery loop from the earliest stages of ideation to implementation and evaluation in real scientific settings.

\subsection*{Data, Materials, and Software Availability} 
Knowledge-graph data are available on Zenodo at \href{https://doi.org/10.5281/zenodo.13900962}{https://doi.org/10.5281/zenodo.13900962} \cite{Scimuse_Zenodo}. Code and evaluation data are available on GitHub at \href{https://github.com/artificial-scientist-lab/SciMuse}{https://github.com/artificial-scientist-lab/SciMuse}.

\subsection*{Ethics Statement} 
The research was reviewed and approved by the Ethics Council of the Max Planck Society.

\subsection*{Acknowledgements}
The authors thank Sören Arlt for organizational help in coordinating the human idea-generation process. The authors are also deeply grateful to all researchers who took the time to participate in our study. In addition, the authors thank the teams behind OpenAlex, arXiv, bioRxiv, medRxiv, and chemRxiv for making scientific resources openly accessible. X.G. acknowledges support from the Alexander von Humboldt Foundation and the Deutsche Forschungsgemeinschaft (DFG, German Research Foundation) – Project-ID 398816777 – SFB 1375. M.K. acknowledges support from the European Research Council (ERC) under the European Union’s Horizon Europe research and innovation programme (ERC-2024-STG, 101165179, ArtDisQ), as well as from the German Research Foundation (DFG) through EXC 2064/1, Project 390727645.
 
\bibliography{ref_url}

\clearpage
\newpage
\section*{Supporting Information}
\subsection*{Prompts used in the study}
\textbf{Prompt for refining the concept list --} We applied GPT-3.5 to refine the concept list and improve its quality. In each round, GPT-3.5 was given a random batch of 10 candidate concepts and instructed to retain only entries corresponding to informative descriptors of research activity, while discarding items that were overly general, unclear, or grammatically incorrect. 

\textbf{Prompt for refining the researcher's concept list --} We refine the extracted concepts with GPT-4 to improve their quality for each researcher, retaining only those that are meaningful and relevant to the scientist’s research direction. This process allows us to build a personalized subgraph in the knowledge graph that reflects each scientist’s research direction. 

\textbf{Prompt for generating research ideas --} We use a single fixed prompt template across all generation conditions. The prompt itself was held constant, and only the source of the concepts varied, namely whether the LLM was given a high-citation-predicted concept pair, a random concept pair, or no externally supplied concept pair (titles-only GPT baseline). The prompt used to generate research ideas from concept pairs in the knowledge graph is shown below. Instead of relying on the knowledge graph to provide ``concept1'' and ``concept2,'' GPT-4 can also be instructed to extract these concepts directly from the research paper titles of Researchers A and B, respectively. These identified concepts can then be used within the same prompt structure to generate ideas.

\begin{figure}[!t]
    \centering
    \includegraphics[width=1\linewidth]{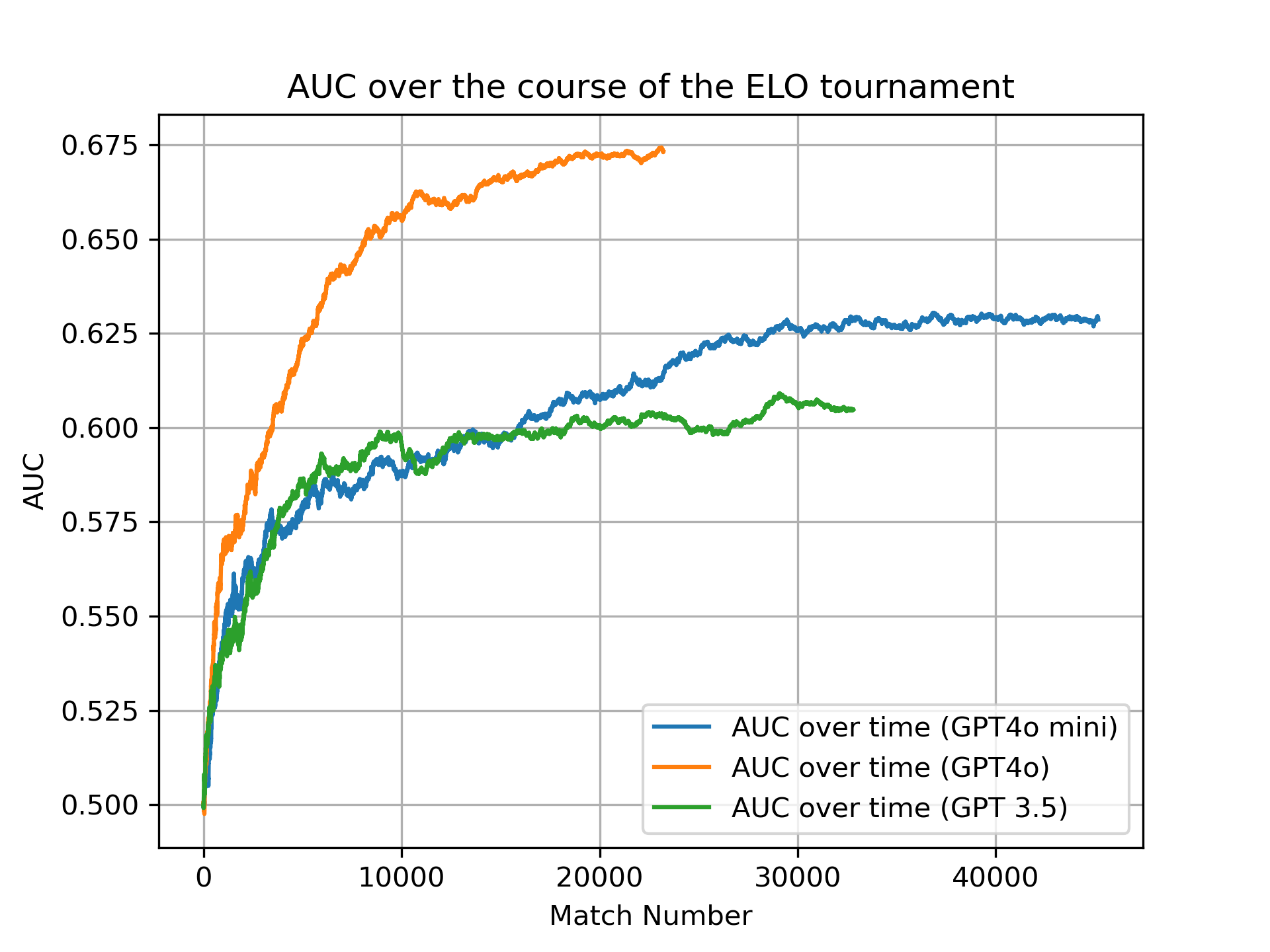}
    \caption{\textbf{Zero-shot ranking of research suggestions by LLMs.} Research suggestions generated using the knowledge graph together with GPT-4 were ranked by GPT-4o, GPT-4o-mini, and GPT-3.5 without using human evaluation data for ranking. Human evaluations were used only afterward to assess ranking quality. The ranking was performed through pairwise comparisons, in which the LLM was asked to select the more interesting of two suggestions given the research background of the evaluating researcher. One match corresponds to one such pairwise comparison. Repeating these pairwise comparisons allows ELO scores to be computed for each generated research idea.}
    \label{fig:ELOtournament}
\end{figure}

\textbf{Prompt for zero-shot ranking --} We ranked 2,996 research suggestions, previously evaluated by human experts, using GPT-3.5, GPT-4o, and GPT-4o-mini. For each pair of randomly selected suggestions, we asked the LLMs to determine which one would be more interesting, taking into account the personalized research interests of the evaluating human expert. This pairwise comparison was repeated between 22,000 and 45,000 times, depending on the model (GPT-4o and GPT-4o-mini, respectively). We treated this procedure as a tournament in which all 2,996 suggestions compete in pairwise matches. Using the Elo ranking system, each suggestion was initialized with a score of 1400. Each GPT-based comparison then updated the Elo scores based on the outcome, producing a final ranking of suggestions from highest to lowest score. We evaluated the quality of this ranking by calculating the AUC, measuring how well the ranked list aligned with the human evaluations of interest (Fig.~\ref{fig:ELOtournament}).  

We have explored manual and automated improvements of the prompts, both for the research question design and the zero-shot prediction. Specifically, we attempted to improve the prompts for the idea generation using GPT-4o. While the prompts were more structured, a small-scale evaluation did not show any improvement in terms of more interesting results.

\subsection*{GPT-4o and GPT-o1 for idea generation}
We conducted two small-scale tests where GPT-4 and GPT-4o generated ideas using the exact same settings described above. In the first test, three research group leaders evaluated 180 pairs of questions (one generated by GPT-4 and the other by GPT-4o using the same prompt). They found 31.1\% of GPT-4 answers to be more interesting, while 60.56\% favored GPT-4o answers (8.3\% were draw). In the second test, a research group leader evaluated 11 pairs of questions (one generated by GPT-4o and the other by GPT-o1 with the same prompt), and all 11 ideas generated by GPT-o1 were ranked as more interesting. These additional small-scale tests suggest that improved models can enhance idea generation, thus directly improving the results of \textsc{SciMuse}.

\subsection*{Human evaluation}
\begin{figure*}[!t]
    \centering
    \includegraphics[width=0.7\linewidth]{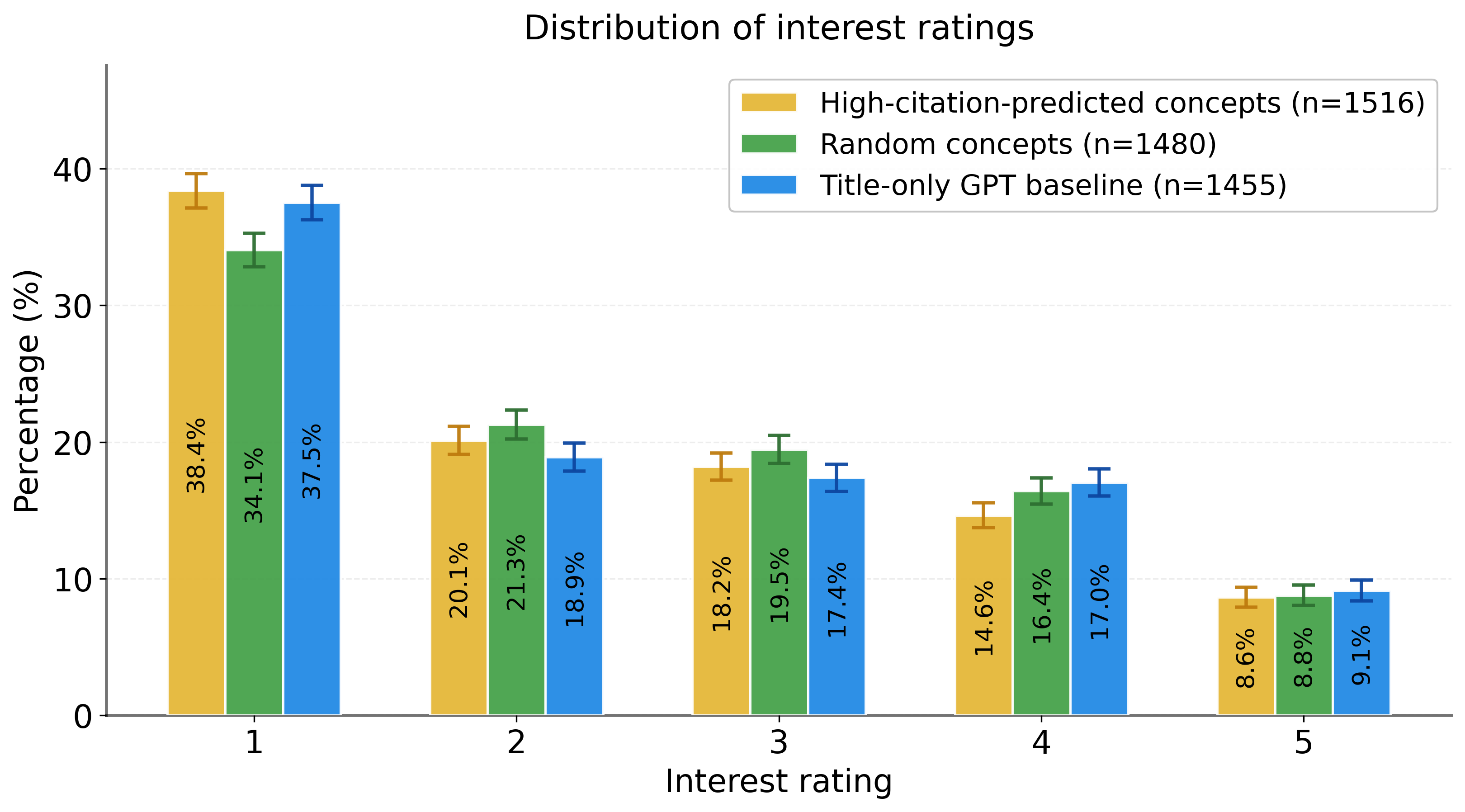}
    \caption{\textbf{Interest levels across different generation methods.} Research ideas are generated using three methods: (1) high-citation-predicted concept pairs (adapting the computational methods from~\cite{gu2025forecasting}) from the researchers' subnetwork ($n=1{,}516$), (2) random concepts from the researchers' subnetwork ($n=1{,}480$), and (3) title-only GPT baseline (i.e., no concepts provided by the knowledge graph) ($n=1{,}455$). Bars show the percentage of ratings falling in each category; error bars denote $\pm$SEM of the proportion.}
    \label{fig:distribution_interest_ratings}
\end{figure*}

\begin{figure*}[!t]
    \centering
    \includegraphics[width=0.95\linewidth]{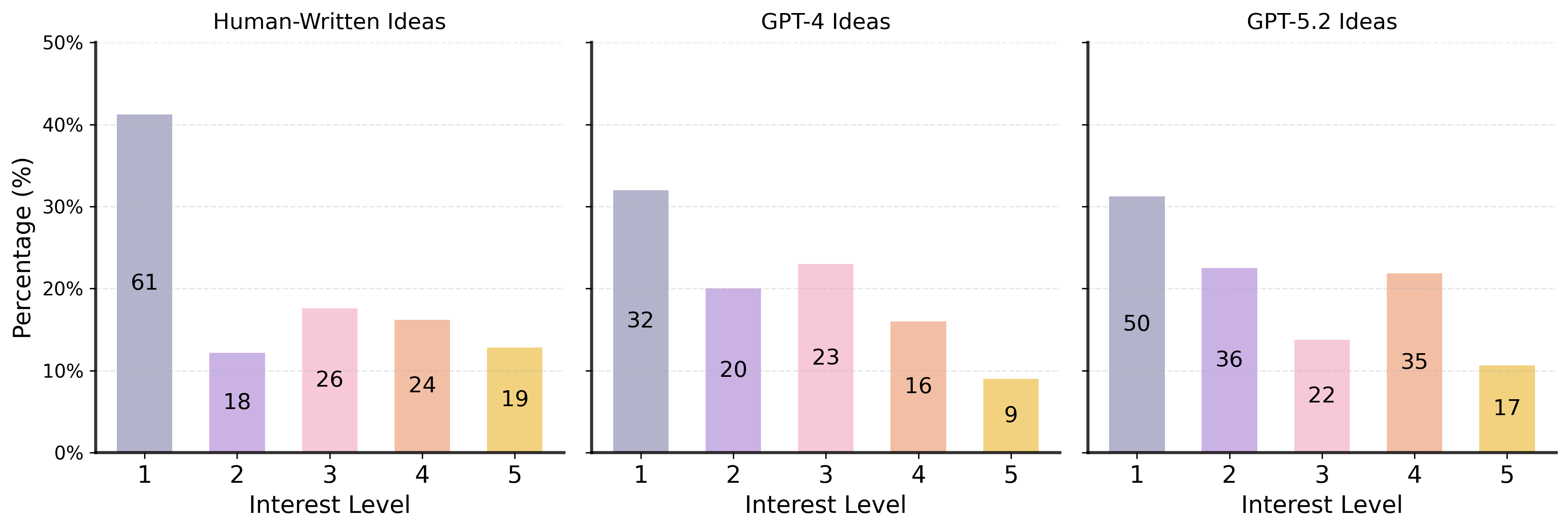}
    \caption{\textbf{Distribution of interest ratings for human-written and AI-generated research ideas.} Each panel shows the percentage of ratings in each interest category (1--5) for one source of ideas, with raw counts annotated inside each bar. The three sources are human-written ideas ($n=148$), GPT-4-generated ideas ($n=100$), and GPT-5.2-generated ideas ($n=160$).}
    \label{fig:distribution_interest_ratings_human_AI}
\end{figure*}

Here we present the interest-level distributions for research suggestions generated using three different methods, based on ratings from 1 (`not interesting') to 5 (`very interesting'). A total of 1516, 1480, and 1455 ratings were collected for the high-citation-predicted, random, and titles-only GPT baseline (no knowledge graph) conditions, respectively. Mean interest ratings were similar across methods: $2.35 \pm 0.04$ (high-citation-predicted concept pair), $2.45 \pm 0.04$ (random concept pair), and $2.41 \pm 0.04$ (titles-only GPT baseline), reported as mean $\pm$ SEM, where $\mathrm{SEM} = s/\sqrt{n}$ with $s$ the sample standard deviation of individual ratings. Likewise, the fraction of ideas receiving a high-interest rating (4 or 5 out of 5) was comparable across conditions: \((23.3 \pm 1.09)\%\) (353/1516) for ideas based on high-citation-predicted concepts, \((25.2 \pm 1.13)\%\) (373/1480) for ideas based on random concepts, and \((26.2 \pm 1.15)\%\) (381/1455) for the titles-only GPT baseline. The uncertainty is the standard error of the mean of each rating-bin proportion. This indicates that interest levels are broadly similar whether or not concepts from the knowledge graph are provided. In Figure~\ref{fig:distribution_interest_ratings}, we show the distribution of interest ratings on a 5-point scale from 1 (`not interesting') to 5 (`very interesting') for the research suggestions generated using three different methods. For a bin containing $k$ ratings out of $n$ total, the sample proportion is $\hat{p} = k/n$, and the SEM is computed as $\mathrm{SEM} = \sqrt{\hat{p}(1-\hat{p})/({n-1})}$. Figure~\ref{fig:distribution_interest_ratings_human_AI} shows the corresponding raw distributions for the human-written baseline ($n=148$), GPT-4-generated ($n=100$), and GPT-5.2-generated ($n=160$) ideas, evaluated by the same six research group leaders under identical conditions. Raw counts are annotated inside each bar. The distributions are broadly comparable across all three sources, consistent with the similar mean scores and high-interest fractions reported in the main text.

\subsection*{Predicting high interest}
\begin{figure*}[!t]
    \centering
    \includegraphics[width=1\linewidth]{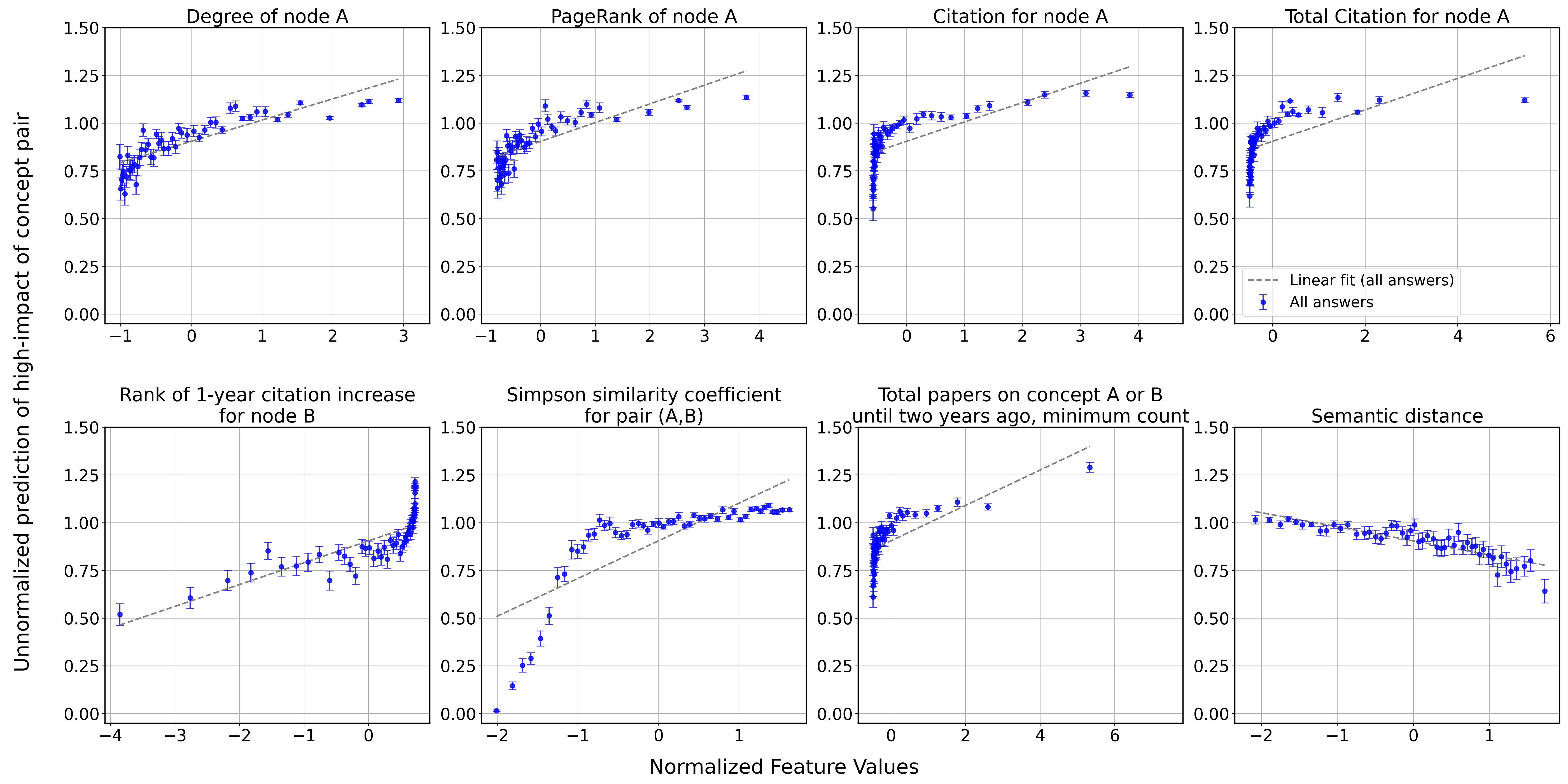}
    \caption{\textbf{Graph Features vs Predicted Citations.} The predicted citation impact has different correlations with knowledge graph features as the empirical interest-values.}
    \label{fig:impact_feature_data}
\end{figure*}

\textcolor{black}{\textbf{Impact-predictions and graph features --}
We find that predicted citation impact, obtained using the neural-network architectures described in \cite{gu2025forecasting}, relates differently to several knowledge-graph features than empirical expert-rated interest (Fig.~\ref{fig:impact_feature_data}). For example, while the expert-rated interest values decreases with the degree, PageRank, citation count, and total citation count of node A, predicted citation impact \textit{increases} with these features. For other features, the two quantities exhibit the same trend; for instance, both the interest value and predicted impact decrease with increasing semantic distance.  A deeper science-of-science investigation of this phenomenon would a direction for future work.}

\textbf{Predicting high interest from knowledge graph features with neural networks --} We aim to predict whether a research proposal will receive a high-interest rating using only information derived from the knowledge graph and excluding the final GPT-generated text. Specifically, human evaluators rated each proposal on a scale from 1 to 5, ranging from \textit{not interesting} to \textit{very interesting}. We train a neural network to predict the interest rating and evaluate its ability to distinguish high-interest proposals, defined as those receiving a rating of 4 or 5, from proposals receiving ratings below 4. We formulate this as a binary classification task.

The input to the neural network model, implemented in PyTorch \cite{paszke2019pytorch}, consists of network-theoretic features extracted from the knowledge graph. For each concept pair associated with a research project, we compute 144 features. The first 141 are adapted from the features used to predict the future impact of concept pairs in \cite{gu2025forecasting}. These include node-level properties, such as node degree and PageRank \cite{page1998pagerank}, and edge-level properties, such as Simpson similarity and the S\o{}rensen--Dice coefficient \cite{barabasi2016network}. Several features also incorporate impact-related information, such as recent citation counts. A detailed description of these 141 features is provided in Ref.~\cite{gu2025forecasting} and in the accompanying Impact4Cast repository: \url{https://github.com/artificial-scientist-lab/Impact4Cast}. The remaining three features consist of the predicted impact and two distance measures between the researchers' subgraphs. The first distance measure is based only on the concepts appearing in the concept lists of Researchers A and B. The second additionally takes into account the broader neighborhood of the two subgraphs by computing semantic distances between neighboring concepts and the concepts contained in the subgraphs. Together, these features serve as input to the neural network for predicting whether a proposal will receive a high-interest rating.

Given the relatively small dataset of 2,996 evaluated suggestions with associated knowledge-graph properties, we use a compact neural network with dropout. For each training set, we evaluate each of the 144 candidate features individually using the ROC AUC for distinguishing ratings of 4 or 5 from ratings below 4. Because both positive and negative associations with high interest may be informative, features are ranked according to the absolute deviation of their AUC from the chance level of 0.5. The 25 highest-ranking features are selected from the training data and used as inputs to the neural network for that run. The network contains one hidden layer with 50 neurons, ReLU activation, and 20\% dropout, followed by a single linear output neuron. The model is trained to predict the original interest ratings using mean squared error as the loss function and the Adam optimizer. To obtain a robust estimate of predictive performance, we use Monte Carlo cross-validation. For each repetition, the dataset is randomly divided into training, validation, and test sets containing 75\%, 15\%, and 10\% of the data, respectively. Feature selection, normalization, and neural-network training are repeated separately for each split. Model training is monitored using the validation loss, and the model parameters corresponding to the lowest validation loss are retained. Performance is then evaluated on the held-out test set. We repeat this procedure for 100 random splits and report the performance averaged across these repetitions.

\begin{figure*}[!t]
    \centering
    \includegraphics[width=1\linewidth]{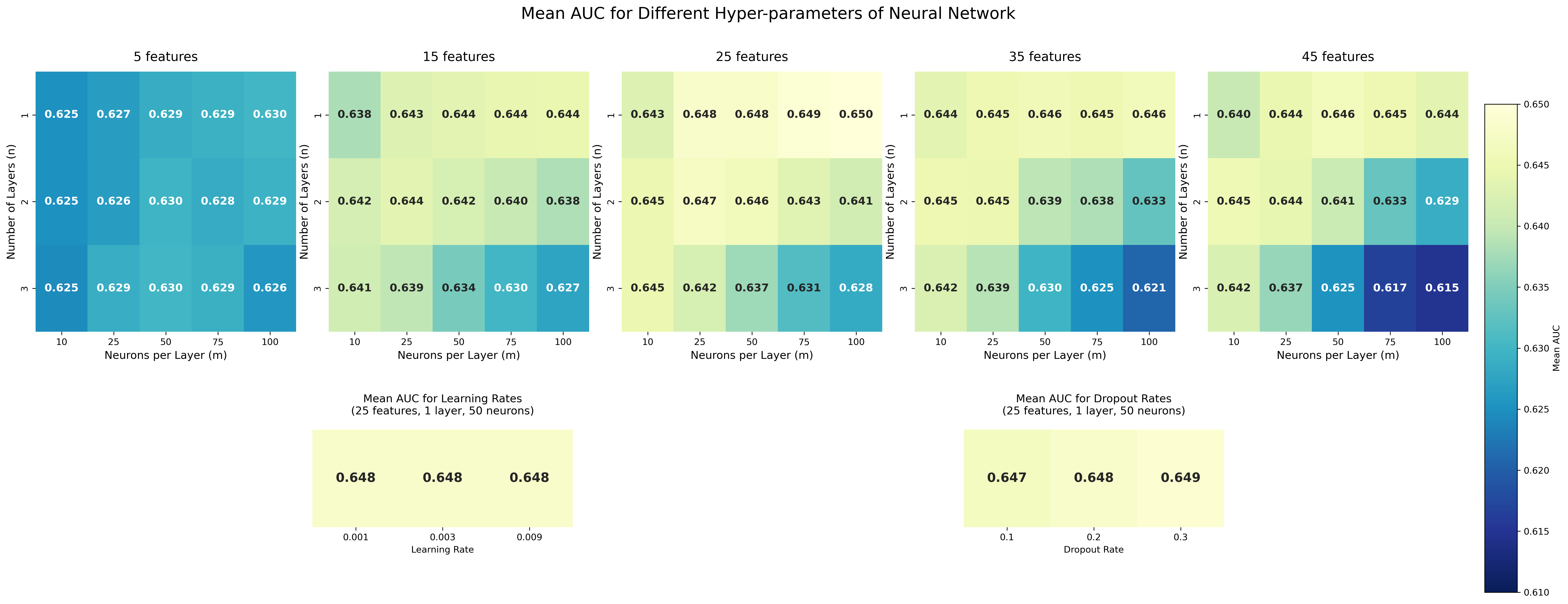}
    \caption{\textbf{Choice of alternative hyper-parameters for training of neural network.} We analyse the prediction of interest level quality (in terms of AUC) with different parameters of the neural network, such as different number of features, different number of layers and neurons, learning rate and drop-out rate. We see that the final results are robust under variations of the hyper-parameters.}
    \label{fig:nn_hyper}
\end{figure*}

We do not perform extensive hyperparameter optimization and instead use a fixed set of reasonable hyperparameter values:  learning rate \(=0.003\), dropout \(=20\%\), weight decay \(=0.0007\), training set \(=75\%\), validation set \(=15\%\), and test set \(=10\%\). In Fig.~\ref{fig:nn_hyper}, we examine alternative hyperparameter settings and find that the results remain robust under different variations. Given the small dataset and compact network architecture, training is computationally inexpensive and can be performed on a standard CPU in less than 15 minutes without parallelization.

\begin{figure*}[!h]
    \centering
    \includegraphics[width=1\linewidth]{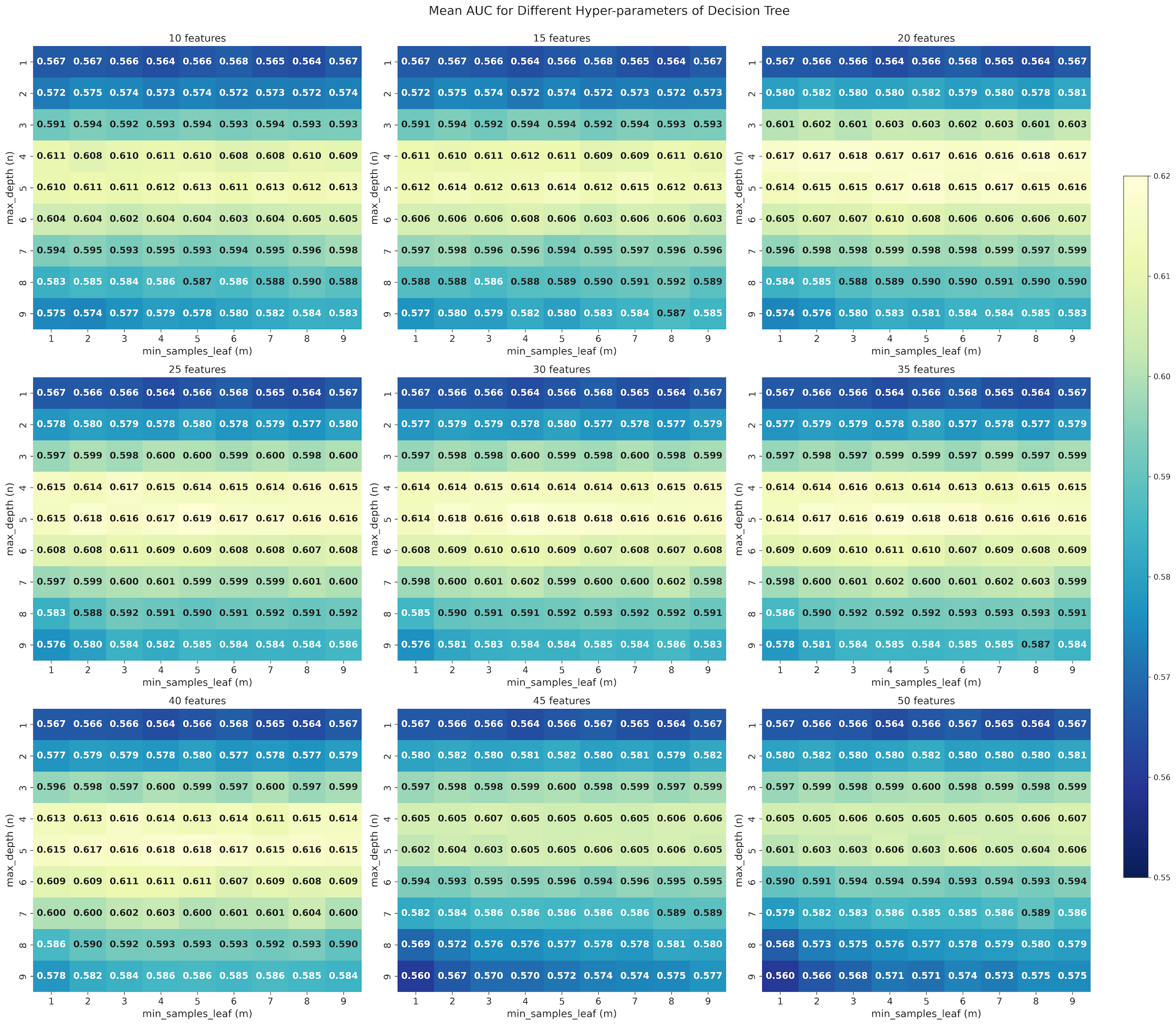}
    \caption{\textbf{Choice of hyper-parameters for training of decision tree.} The model is trained using Monte Carlo cross-validation until a statistical uncertainty of $\sigma$=0.001 is reached. We find that no setting of number of features, maximum depth and minimal sample leaf can reach the performance of the data-efficient neural network.}
    \label{fig:dt_hyper}
\end{figure*}

\begin{figure*}[!h]
    \centering
    \includegraphics[width=1\linewidth]{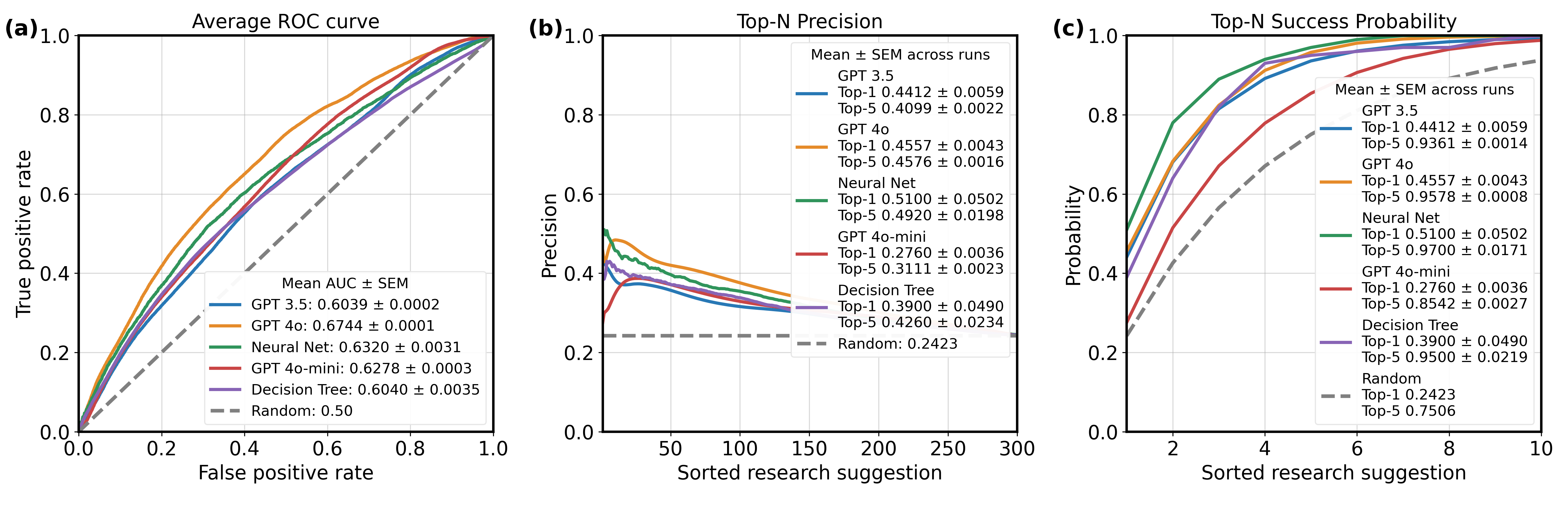}
    \caption{\textbf{Interest predictions with five methods.} We reproduce the results in the main text, and add results from a supervised decision tree training, and an unsupervised GPT4o-mini.}
    \label{fig:all_data}
\end{figure*}

\textbf{Predicting high interest from knowledge graph features with decision trees --} We also experimented with other data-efficient learning methods, in particular decision trees \cite{breiman2017classification} implemented with scikit-learn \cite{scikit-learn}. However, decision trees did not outperform the neural network, as shown in Fig.~\ref{fig:dt_hyper}. These results further support the use of neural networks for supervised ranking of research ideas by their predicted interest, consistent with the comparison shown in Fig.~\ref{fig:all_data}.

\textbf{Prediction of interest with different methods --} Figure~\ref{fig:all_data} summarizes the results for all five methods: supervised prediction using neural networks and decision trees, and zero-shot ranking using GPT-3.5, GPT-4o, and GPT-4o-mini. For each method, we report the AUC, top-\(N\) precision, and top-\(N\) success probability. We find that the neural network outperforms decision trees in the supervised setting, while GPT-4o performs better than other models when ranking is carried out in a zero-shot manner without access to human evaluation data.
\clearpage
\newpage
\begin{tcolorbox}[colframe=black, colback=white, title={Prompt to Refine the Concept List}, label={box:prompt_concept_KG}, before skip=5pt, after skip=5pt]
I am creating a list of scientific concepts, the list has 368,825 entries. I want to cut it down at least 30\%. Can you help me with that? I give you 10 scientific concepts (as a python list called {\ttfamily original\_concepts}), and you return a list {\ttfamily gpt\_list} that only contains meaningful concepts. By meaningful i mean: Each concept should be an informative way to describe someones research. Concepts that are too general, unclear or grammatically incorrect should be removed. Remember i want to remove about a third of the concepts in total, so try to remove about a third of the concept in the list. The new list should be called {\ttfamily gpt\_list}.\\

Just output the python list, nothing else. No other statements or explanations, the full reply should be one python list {\ttfamily gpt\_list}! no print or no comment or anything else.
\label{box:prompt_concept_kgraph} 
\end{tcolorbox}

\begin{tcolorbox}[colframe=black, colback=white, title={Prompt to Refine the Researchers' Concept List}, label={box:prompt_concept},before skip=5pt, after skip=5pt]
A scientist has written the following papers: \\
0) title1 \\
1) title2 \\
2) title3 \\
...\\

I have a noisy list of the researchers' topics of interest, and I would like your help in filtering them. Please look at the list below and return all concepts that are relevant to the scientist's research (based on their paper titles) and meaningful in the context of their research direction. The concepts can be detailed; I mainly want you to filter out concepts that are not meaningful, words that are not concepts, or concepts that are too general for the direction of the scientist (e.g., ``artificial intelligence" might be a meaningful concept for a geologist, but not for a machine learning researcher). Do not change or add any concepts -- only remove or keep them.

\vspace{0.3cm} 
concept list=[c1, c2, c3, c4, c5, c6, ...]
\label{box:prompt_concept} 
\end{tcolorbox}

\begin{tcolorbox}[colframe=black, colback=white, title={Prompt to GPT-4 for Project Idea Generation}, before skip=15pt, after skip=15pt]
Two researchers A and B, with expertise in ``concept1" and ``concept2" respectively, are eager to collaborate on a novel interdisciplinary project that leverages their unique strengths and creates synergy between their fields.\\

To better understand their backgrounds, here are the titles of recent publications from each researcher:\\
Researcher A:\\
1: title1\\
2: title2\\
3: title3\\
...\\
Researcher B:\\
1: title1\\
2: title2\\
3: title3\\
...\\

Please suggest a creative and surprising scientific project that combines ``concept1" and ``concept2". In your response, follow this outline:\\

First, explain ``concept1" and ``concept2" in one short sentence each.\\

Then, do the following three steps 3 times, improving in each time the response:\\
A) Describe 4 interesting and new scientific contexts, in which those two concepts might appear together in a natural and useful way.\\
B) Criticize the 4 contexts (one short sentence each), based on how well the contexts merge the idea of the two concepts.\\
C) Give a 2 sentence summary of your reflections above, on how well one can combine these concepts naturally and interestingly.\\

Then, start finding a project. Taking your reflections from (A-C) into account, define in your response a project title, followed by a brief explanation of the project's main objective.\\

Finally, address the following questions (Take the full reflections (A-C) into account):\\
What specific interesting research questions will this project address, that will lead to innovative novel results? [2 bullet points, one sentence each]
\label{box:prompt_idea} 
\end{tcolorbox}

\begin{tcolorbox}[colframe=black, colback=white, title={Prompt for Zero-Shot Ranking of Research Ideas}, label={box:prompt_zeroshot}, before skip=10pt, after skip=10pt]
I will present two research ideas. The first idea is for Researchers A1 and B1, and the second idea is for Researchers A2 and B2.\\
Researchers A1 and A2 will evaluate how interesting they find the respective ideas.\\
You will determine which of the two suggestions will be considered more interesting.\\\\
The suggestions are randomly ordered, and you should evaluate each suggestion independently and without bias.\\\\
\#\#\# Researcher A1 Context and Suggestion 1:\\
Here are a few papers of Researcher A1:\\
(papersA1)\\\\
Suggestion 1: [suggestion1]\\\\
**Summary for Researcher A1**: Provide a one-sentence summary of Suggestion 1 in the context of Researcher A1.\\\\
\#\#\# Researcher A2 Context and Suggestion 2:\\
Here are a few papers of Researcher A2:\\
(papersA2)\\\\
Suggestion 2: [suggestion2]\\\\
**Summary for Researcher A2**: Provide a one-sentence summary of Suggestion 2 in the context of Researcher A2.\\\\
\#\#\# Evaluation:\\
Based on the summaries and the research interests of A1 and A2, evaluate which suggestion is more likely to be ranked higher in terms of interest.\\
**Result**: If Suggestion 1 is ranked higher by Researcher A1 than Suggestion 2 is by Researcher A2, write 'RESULT: SUGGESTION 1'. Otherwise, write 'RESULT: SUGGESTION 2'.\\
Remember, the suggestions are randomly ordered, and your evaluation should be impartial and based solely on the research interests of A1 and A2.
\end{tcolorbox}

\clearpage
\newpage
\begin{table*}[!h]
\centering
\caption{\textbf{Roles and execution dates of the OpenAI models used in the study.} Note: Model use occurred at different stages of the study. Personalized research suggestions were prepared on a rolling basis after each research group leader agreed to participate, resulting in an extended period of GPT-4 use for generating researcher suggestions.}
\label{table:model_roles}
\begin{tabular}{c|l|l}
\hline
\textbf{Model} & \textbf{Role} & \textbf{Period of use} \\
\hline

GPT-3.5 &
Refinement of the concept list. &
08.2023 \\
\hline

GPT-4 &
Refinement of researcher-specific concept lists. &
09.2023--02.2024 \\
\hline

GPT-4 &
Generation of the main personalized research ideas. &
09.2023--02.2024 \\
\hline

GPT-4o &
Cross-checking of institute classifications. &
09.2024 \\
\hline

GPT-3.5 &
Zero-shot pairwise ranking of research ideas to predict expert-rated interest. &
09.2024 \\
\hline

GPT-4o &
Zero-shot pairwise ranking of research ideas to predict expert-rated interest. &
09.2024 \\
\hline

GPT-4o-mini &
Additional zero-shot ranking analysis reported in the Supporting Information. &
09.2024 \\
\hline

o1 &
Small-scale idea-generation comparison with GPT-4o reported in the Supporting Information. &
09.2024 \\
\hline

GPT-5.2 &
Generation of 160 additional personalized ideas for the human--AI baseline comparison. &
02.2026 \\
\hline
\end{tabular}
\end{table*}



\end{document}